\pdfoutput=1
\documentclass[11pt]{article}

\usepackage[]{naacl2021}

\usepackage{times}
\usepackage{hyperref}
\usepackage{latexsym}
\usepackage{multirow}
\usepackage{bm}
\usepackage{amsmath}
\usepackage{graphicx}
\usepackage{subfig}

\usepackage{ragged2e}
\usepackage{float}

\usepackage{algorithm}
\usepackage{algorithmicx}
\usepackage{algpseudocode}

\algnewcommand\algorithmicinput{\textbf{Input:}}
\algnewcommand\algorithmicoutput{\textbf{Output:}}
\algnewcommand\algorithmicparameter{\textbf{Parameters:}}
\algnewcommand\INPUT{\item[\algorithmicinput]}
\algnewcommand\OUTPUT{\item[\algorithmicoutput]}
\algnewcommand\PARAMETER{\item[\algorithmicparameter]}

\usepackage[T1]{fontenc}
\usepackage[utf8]{inputenc}

\usepackage{microtype}

\title{Empirical Evaluation of Pre-trained Transformers for Human-Level NLP:\\ The Role of Sample Size and Dimensionality}

\author{Adithya V Ganesan, Matthew Matero, Aravind Reddy Ravula,\\ {\bf Huy Vu, \and H. Andrew Schwartz} \\
Department of Computer Science, Stony Brook University \\
\texttt{\{avirinchipur, mmatero, aravula, hvu, has\}@cs.stonybrook.edu}}
\begin{document}
\maketitle
\begin{abstract}

% Human-level NLP tasks, such as predicting mental health, personality, or demographics, usually rely on small sample sizes due to the limited availability of labeled data. 
% In fact, the number of observations is often smaller than the standard 768+ hidden state sizes of each layer within modern transformer-based language models, limiting the ability to effectively leverage transformers. 
In human-level NLP tasks, such as predicting mental health, personality, or demographics, the number of observations is often smaller than the standard 768+ hidden state sizes of each layer within modern transformer-based language models, limiting the ability to effectively leverage transformers.
Here, we provide a systematic study on the role of dimension reduction methods (principal components analysis, factorization techniques, or multi-layer auto-encoders) as well as the dimensionality of embedding vectors and sample sizes as a function of predictive performance.
We first find that fine-tuning large models with a limited amount of data pose a significant difficulty which can be overcome with a pre-trained dimension reduction regime. 
RoBERTa consistently achieves top performance in human-level tasks, with PCA giving benefit over other reduction methods in better handling users that write longer texts.
% We find that auto-encoder style language models (BERT, RoBERTa) often perform better for these human-level tasks than auto-regressive (XLNet, GPT-2), with RoBERTa most consistently achieving top performance.
Finally, we observe that a majority of the tasks achieve results comparable to the best performance with just $\frac{1}{12}$ of the embedding dimensions. 
\end{abstract}

\section{Introduction}
%Transformer~\citep{vaswani2017attention} based language models (LM) have been growing both ways with increase in the number of transformer layers as well as the hidden representation size. 
Transformer based language models (LMs) have quickly become the foundation for accurately approaching many tasks in natural language processing~\cite{vaswani2017attention,devlin-etal-2019-bert}.
Owing to their success is their ability to capture both syntactic and semantic information~\citep{tenney-etal-2019-bert}, modeled over large, deep attention-based  networks (transformers) with hidden state sizes on the order of 1000 over 10s of layers~\cite{liu2019roberta,gururangan-etal-2020-dont}. 
In total such models typically have from hundreds of millions~\citep{devlin-etal-2019-bert} to a few billion parameters~\citep{raffel2020exploring}.
However, the size of such models presents a challenge for tasks involving small numbers of observations, such as for the growing number of tasks focused on human-level NLP.   

% \begin{table}[]
%     \centering
%     \begin{tabular}{llcc}
%     \hline
%     \bf Task& \bf Metric& \bf 768 dims& \bf 128 dims \\
%     \hline
%     Age &  Pearson-r & $0.649_{\ 0.04}$& $\textbf{0.650}_{\ 0.03}$\\
%     Ext &  Pearson-r$^*$& $0.123_{\ 0.06}$& $\textbf{0.139}_{\ 0.06}$\\
%     Ope &  Pearson-r$^*$& $0.157_{\ 0.04}$& $\textbf{0.182}_{\ 0.03}$\\
%     BSAG&  Pearson-r$^*$& $0.325_{\ 0.04}$& $\textbf{0.342}_{\ 0.03}$\\
%     \hline
%     \end{tabular}
%     \caption{Reducing contextual embeddings to lower dimensions using PCA consistently improves the performance with small sample sizes. The table compares the performance of RoBERTa-base features (768 dims) and PCA reduced features (128 dims) with 100 training samples averaged over 10 runs. Pearson-r$^*$ indicates dis-attenuated Pearson-r and the numbers in subscript denote the standard deviation.}
%     \label{tab:intro}
% \end{table}
 
Human-level NLP tasks, rooted in computational social science, focus on making predictions about people from their language use patterns. 
Some of the more common tasks include age and gender prediction~\cite{sap2014developing,morgan2017predicting}
, personality~\cite{Park2015AutomaticPA,lynn-etal-2020-hierarchical}, and mental health prediction~\cite{coppersmith-etal-2014-quantifying,guntuku2017detecting,lynn-etal-2018-clpsych}.
Such tasks present an interesting challenge for the NLP community to model the people behind the language rather than the language itself, and the social scientific community has begun to see success of such approaches as an alternative or supplement to standard psychological assessment techniques like questionnaires
~\cite{kern2016gaining,Eichstaedt2018FacebookLP}.
Generally, such work is helping to embed NLP in a greater social and human context~\cite{hovy2016social,lynn2019tweet}.

Despite the simultaneous growth of both (1) the use of transformers and (2) human-level NLP, the effective merging of transformers for human-level tasks has received little attention.  
In a recent human-level shared task on mental health, most participants did not utilize transformers~\cite{zirikly-etal-2019-clpsych}. 
A central challenge for their utilization in such scenarios is that the number of training examples (i.e. sample size) is often only hundreds while the parameters for such deep models are in the hundreds of millions. 
For example, recent human-level NLP shared tasks focused on mental health have had $N = 947$~\cite{milne-etal-2016-clpsych}, $N = 9,146$~\cite{lynn-etal-2018-clpsych} and $N = 993$~\cite{zirikly-etal-2019-clpsych} training examples. 
Such sizes all but rules out the increasingly popular approach of fine-tuning transformers whereby all its millions of parameters are allowed to be updated toward the specific task one is trying to achieve~\cite{devlin-etal-2019-bert,mayfield-black-2020-fine}.
Recent research not only highlights the difficulty in fine-tuning with few samples~\citep{jiang-etal-2020-smart} but it also becomes unreliable even with thousands of training examples~\cite{mosbach2020stability}.

%\has{if time, look for a better citation for this last point; the paper cited has only been published on arxiv with only 1 citation thus far}. Nothing that fits well, still looking

On the other hand, some of the common transformer-based approaches of deriving contextual embeddings from the top layers of a \textit{pre-trained} model~\cite{devlin-etal-2019-bert, clark2019boolq} still leaves one with approximately an equal number of embedding dimensions as training size.
In fact, in one of the few successful cases of using transformers for a human-level task, further dimensionality reduction was used to avoid overfit~\cite{matero-etal-2019-suicide}, but an empirical understanding of the application of transformers for human-level tasks --- which models are best and the  relationship between embedding dimensions, sample size, and accuracy --- has yet to be established. 

%This dearth stems from the limited availability of labelled data for these tasks and issues around people agreeing to share their data.
%At the same time that transformers have grown within NLP and AI communities, other fields have begun to show promising new  new applications of NLP to the social and psychological sciences have emerged

In this work, we empirically explore strategies to effectively utilize transformer-based LMs for relatively small sample-size human-level tasks. We provide the first systematic comparison of the most widely used transformer models for demographic, personality, and mental health prediction tasks. 
Then, we consider the role of dimension reduction to address the challenge of applying such models on small sample sizes, yielding a suggested minimum number of dimensions necessary given a sample size for each of demographic, personality, and mental health tasks\footnote{dimension reduction techniques can also be pre-trained leveraging larger sets of unlabeled data}.
While it is suspected that transformer LMs contain more dimensions than necessary for document- or word-level NLP~\cite{li-eisner-2019-specializing,bao-qiao-2019-transfer}, this represents the first study on transformer dimensionality for human-level tasks. 
%The contributions of this work include: (1) systematic comparison of the most widely used transformer-based LMs for human-level tasks, (2) we show that while fine-tuning is mostly elusive at smaller sample sizes ($ < 500$), dimensionality reduction techniques,\footnotes{dimension reduction techniques can also be pre-trained leveraging larger sets of unlabeled data} can improve prediction accuracies, and (3) understanding the effect of the number of dimensions on performance as a function of train sample size.

\section{Related Work}
Recently, NLP has taken to human-level predictive tasks using increasingly sophisticated techniques. 
The most common approaches use n-grams and LDA~\cite{blei2003latent} to model a person's language and behaviors~\cite{resnik-etal-2013-using,kern2016gaining}. 
Other approaches utilize word embeddings~\cite{mikolov2013distributed, pennington-etal-2014-glove} and more recently, contextual word representations~\cite{ambalavanan-etal-2019-using}. 

Our work is inspired by one of the top performing systems at a recent mental health prediction shared task~\cite{zirikly-etal-2019-clpsych} that utilized transformer-based contextualized word embeddings fed through a non-negative matrix factorization to reduce dimensionality~\cite{matero-etal-2019-suicide}. 
While the approach seems reasonable for addressing the dimensionality challenge in using transformers, many critical questions remain unanswered: 
(a) Which type of transformer model is best? (b) Would fine-tuning have worked instead? and (c) Does such an approach generalize to other human-level tasks?
Most of the time, one does not have a luxury of a shared task for their problem at hand to determine a best approach. 
Here, we look across many human-level tasks, some of which with the luxury of having relatively large sample sizes (in the thousands) from which to establish upper-bounds, and ultimately to draw generalizable information on how to approach a human-level task given its domain (demographic, personality, mental health) and sample size. 

Our work also falls in line with a rising trend in AI and NLP to quantify the number of dimensions necessary.
While this has not been considered for human-level tasks, it has been explored in other domains. 
The post processing algorithm~\citep{mu2018all} of the static word embeddings motivated by the power law distribution of maximum explained variance and the domination of mean vector turned out to be very effective in making these embeddings more discriminative. 
% Principal component analysis~\citep{tipping1999probabilistic} in conjunction to this algorithm demonstrated the improved performance of embeddings with fewer dimensions~\citep{raunak-etal-2019-effective}. 
The analysis of contextual embedding models~\citep{ethayarajh-2019-contextual} suggest that the static embeddings contribute to less than 5\% to the explained variance, the contribution of the mean vector starts dominating when contextual embedding models are used for human-level tasks. This is an effect of averaging the message embeddings to form user representations in human-level tasks. This further motivates the need to process these contextual embeddings into more discriminative features.

Lastly, our work weighs into the discussion on just which type of model is best in order to produce effective contextual embedding models. 
%These models fall into 5 broad categories: (1) auto-encoders, (2) auto-regressive, (3) sequence-to-sequence, (4) multi-modal, and (5) retrieval based. However, a majority of models fall under the first two categories. 
A majority of the models fall under two broad categories based on how they are pre-trained - auto-encoders (AE) and auto-regressive (AR) models. We compare the performance of AE and AR style LMs by comparing the performance of two widely used models from each category with comparable number of parameters. From the experiments involving BERT, RoBERTa~\citep{liu2019roberta}, XLNet~\cite{yang2019xlnet} and GPT-2~\citep{radford2019language}, we find that AE based models perform better than AR style models (\textit{with comparable model sizes}), and RoBERTa is the best choice amongst these four widely used models.

\section{Data \& Tasks}
We evaluate approaches over 7 human-level tasks spanning Demographics, Mental Health, and personality prediction. The 3 datasets used for these tasks are described below.

\paragraph{FB-Demogs. (age, gen, ope, ext)} One of our goals was to leverage one of the largest human-level datasets in order to evaluate over subsamples of sizes. 
For this, we used the Facebook demographic and personality dataset of~\citet{Kosinski2013PrivateTA}. 
The data was collected from approximately 71k consenting participants who shared Facebook posts along with demographic and personality scores from Jan-2009 through Oct-2011. 
The users in this sample had written at least a 1000 words and had selected English as their primary language. Age (age) was self-reported and limited to those 65 years or younger (data beyond this age becomes very sparse) as in~\citep{sap2014developing}. 
Gender (gen) was only provided as a limited single binary, male-female classification. 
% Since we are interested in relative differences in performance and are exploring many large models, we limit each user to at most 20 messages (approximately the median number of messages), randomly sampled, to save compute time. 

Personality was derived from the Big 5 personality traits questionnaires, including both extraversion (ext - one's tendency to be energized by social interaction) and openess (ope, one's tendency to be open to new ideas)~\citep{schwartz2013personality}. 
Disattenuated Pearson correlation\footnote{Disattenuated Pearson correlation helps account for the error of the measurement instrument~\citep{Kosinski2013PrivateTA, Murphy1988PsychologicalTP}. Following~\citep{lynn-etal-2020-hierarchical}, we use reliabilities: $r_{xx}= 0.70$ and $r_{yy}= 0.77$.} ($r_{dis}$) was used to measure the performance of these two personality prediction tasks.

\paragraph{CLPsych-2018. (bsag, gen2)} The CLPsych 2018 shared task~\citep{lynn-etal-2018-clpsych} consisted of sub-tasks aimed at early prediction of mental health scores (depression, anxiety and BSAG\footnote{Bristol Social Adjustment Guide~\citep{ghodsian1977children} scores contains twelve sub-scales that measures different aspects of childhood behavior.} score) based on their language. The data for this shared task~\citep{ncds} comprised of English essays written by 11 year old students along with their gender (gen2) and income classes. There were 9217 students' essays for training and 1000 for testing. The average word count in an essay was less than 200. Each essay was annotated with the student's psychological health measure, BSAG (when 11 years old) and distress scores at ages 23, 33, 42 and 50. 
%The gender and income class of the students were also provided.
This task used a disattenuated pearson correlation as the metric ($r_{dis}$).

\paragraph{CLPsych-2019. (sui)} This 2019 shared task~\citep{zirikly-etal-2019-clpsych} comprised of 3 sub-tasks for predicting the suicide risk level in reddit users. This included  a history of user posts on \texttt{r/SuicideWatch} (SW), a subreddit dedicated to those wanting to seek outside help for processing their current state of emotions. Their posts on other subreddits (NonSuicideWatch) were also collected. The users were annotated with one of the 4 risk levels: none, low, moderate and severe risk based on their history of posts. In total this task spans 496 users in training and 125 in testing.
We focused on Task A, predicting suicide risk of a user by evaluating their (English) posts across SW, measured via macro-F1.

\begin{table}[tbh]
    \centering
    \begin{small}
    \begin{tabular}{|l|c|c|c|}
    %\begin{tabular}{|p{0.035\textwidth}|p{0.12\textwidth}|p{0.12\textwidth}|p{0.12\textwidth}|}
        \hline
        &
        FB-Demogs&
        \begin{tabular}{@{}c@{}}
        CLPsych \\ 2018
        \end{tabular}&
        \begin{tabular}{@{}c@{}}
        CLPsych \\ 2019
        \end{tabular}
        \\
        
        \hline
        &~\citeauthor{sap2014developing}&~\citeauthor{lynn-etal-2018-clpsych}&~\citeauthor{zirikly-etal-2019-clpsych}\\ 
        %\cline{2-4}
        $N_{pt}$ & 56,764& 9,217& 496\\
        $N_{max}$& 10,000& 9,217& 496\\
        $N_{te}$& 5,000& 1,000& 125\\
        \hline

    \end{tabular}
    \end{small}
    \caption{ \textit{Summary of the datasets.} $\bm{N_{pt}}$ is the number of users available for pre-training the dimension reduction model; $\bm{N_{max}}$ is the maximum number of users available for task training. For CLPsych 2018 and CLPsych 2019, this would be the same sample as pre-training data. For Facebook, a disjoint set of 10k users was available for task training; $\bm{N_{te}}$ is the number of test users. This is always a disjoint set of users from the pre-training and task training samples.} 
    \label{tab:data}
\end{table}

% \begin{table}[!htb]
%     \centering
%     \begin{tabular}{|llc|}
%          \hline
%          \textbf{Task} & \textbf{Dataset}& \textbf{Metric}\\
%          \hline
%          \multicolumn{3}{|l|}{\textbf{Demographic Prediction}} \\
%          Age \textit{(age)}& Facebook& Pearson-r\\
%          Gender \textit{(gen)}& Facebook& Macro-F1\\
%          Gender \textit{(gen2)}& CLPsych '18& Macro-F1\\
%          \hline
%          \multicolumn{3}{|l|}{\textbf{Personality Prediction}} \\
%          Openness \textit{(ope)}& Facebook& $r_{dis}$\protect \footnotemark\\
%          Extraversion \textit{(ext)}& Facebook& $r_{dis}$\\
%          \hline
%         \multicolumn{3}{|l|}{\textbf{Mental Health Prediction}} \\
%          \begin{tabular}[c]{@{}l@{}}Childhood\\ behavior \textit{(bsag)}\end{tabular}& CLPsych '18& $r_{dis}$\\
%          Suicide risk \textit{(sui)}& CLPsych '19& Macro-F1\\
%          \hline
%     \end{tabular}
%     \caption{\has{would remove this. make sure it's in text; doesn't need table} The tasks from the respective datasets used for the experiments and the metric chosen to evaluate them. We pick 3 demographic prediction tasks, 2 personality prediction and 2 mental health prediction tasks. }
%     \label{tab:tasks}
% \end{table}

% \footnotetext{Disattenuated Pearson correlation helps account for the error of the measurement instrument~\citep{Kosinski2013PrivateTA, Murphy1988PsychologicalTP}. Following~\citep{lynn-etal-2020-hierarchical}, we use reliabilities: $r_{xx}= 0.70$ and $r_{yy}= 0.77$.}

\section{Methods}

Here we discuss how we utilized representations from transformers, our approaches to dimensionality reduction, and our technique for robust evaluation using bootstrapped sampling. 

\subsection{Transformer Representations}
% \paragraph{Transformer Representations. }
The second to last layer representation of all the messages was averaged to produce a 768 dimensional feature for each user\footnote{The second to last layer was chosen owing to its consistent performance in capturing semantic and syntactic structures~\cite{jawahar-etal-2019-bert}.}. These user representations are reduced to lower dimensions as described in the following paragraphs. The message representation from a layer was attained by averaging the token embeddings of that layer.
To consider a variety of transformer LM architectures, we explored two popular auto-encoder (BERT and RoBERTa) and two auto-regressive (XLNet and GPT-2) transformer-based models. 

For fine-tuning evaluations, we used the transformer based model that performs best across the majority of our task suite. Transformers are typically trained on single messages or pairs of messages, at a time. Since we are tuning towards a human-level task, we label each user's message with their human-level attribute and treat it as a standard document-level task~\citep{morales-etal-2019-investigation}. Since we are interested in relative differences in performance, we limit each user to at most 20 messages - approximately the median number of messages, randomly sampled, to save compute time for the fine tuning experiments.

\begin{algorithm}%[!thb]
\caption{Dimension Reduction and Evaluation}
\label{alg:method}

\begin{algorithmic}[1]
\item[\textbf{Notation:}] $h_D$: hidden size, $ f(\cdot)$: function to train dimension reduction, $\theta$: Linear Model,  $g(\cdot, \cdot)$: Logistic loss function for classification and L2 loss for regression, $\eta$: learning rate, $T$: Number of iterations (100).

\item[\textbf{Data:}] $D_{pt}\in R^{N_{pt}\times h_D}$: Pre-training embeddings, $D_{max} \in  R^{N_{max}\times h_D}$: Task training embeddings, $D_{te} \in  R^{N_{te}\times h_D}$: Test embeddings, $Y_{max}$: Outcome for train set, $Y_{te}$: Outcome for test set.

  \State $W \leftarrow f(D_{pt})$
  
  \State $\Bar D_{max} \leftarrow D_{max}W$
  \State $\Bar D_{te} \leftarrow D_{te}W$
  
  \For{$i=1,\hdots,10$}
  \State $\theta_i^{(0)} \leftarrow \vec 0$
  \State Sample $(\Bar D_{ta}, Y_{ta})$ from $(\Bar D_{max}, Y_{max})$
  \For{$j=1,\hdots,T$}
  \State $\theta_i^{(j)} \leftarrow \theta_i^{(j-1)} - \eta \nabla g(\Bar D_{ta}, Y_{ta})$
  \EndFor
  \State $\Hat Y_{te_i} \leftarrow \Bar D_{te} \theta_i^{(T)}$
  \EndFor
  
  \State $Evaluate(\Hat Y_{te}, Y_{te})$ 

\end{algorithmic}
\end{algorithm}

% \has{cite someone who has done this} 
%HAS: this is evaluation - We perform fine-tuning across 2 tasks (Age and Gender) to evaluate the viability of applying this common NLP technique to small human-level datasets, in this case a sub-sampled dataset of 100 and 500 Facebook users. % In our fine-tuning configuration we freeze all but the top 2 layers of the best LM, to prevent over fitting and vanishing gradients at the lower layers~\cite{mosbach2020stability}. 
% We also apply early stopping (patience=3), L2-regularization (in the form of weight-decay on SGD optimizer, set to .01), and dropout (between .05 and .1 depending on task).

\subsection{Dimension Reduction}
% \paragraph{Dimension Reduction. }
We explore singular value decomposition-based methods such as Principal components analysis (PCA)~\citep{halko2011finding}, Non-negative matrix factorization (NMF)~\citep{fevotte2011algorithms} and Factor analysis (FA) as well as a deep learning approach: multi-layer non linear auto encoders (NLAE)~\citep{hinton2006reducing}.
We also considered the post processing algorithm (PPA) of word embeddings\footnote{The 'D' value was set to $\lfloor{\frac{number\:of\:dimensions}{100}}\rfloor$.}~\citep{mu2018all} that has shown effectiveness with PCA on word level~\citep{raunak-etal-2019-effective}.
Importantly, besides transformer LMs being pre-trained, so too can dimension reduction. Therefore, we distinguish: (1) learning the transformation from higher dimension to lower dimensions (preferably on a large data sample from the same domain) and (2) applying the learned transformation (on the task's train/test set). 
For the first step, we used a separate set of ~56k unlabeled user data in the case of FB-demog\footnote{these pre-trained dimension reduction models are made available.}. For CLPsych-2018 and -2019 (where separate data from the exact domains was not readily available), we used the task training data to train the dimension reduction. 
Since variance explained in factor analysis typically follows a power law, these methods transformed the 768 original embedding dimensions down to $k$, in powers of 2: 16, 32, 64, 128, 256 or 512. 

% \subsection{Evaluation}
% The performance of each model trained on a bootstrapped sample is recorded on the test set. This set of 10 scores for each $N_{ta}$ value is reduced to the mean and standard deviation. For each $N_{ta}$ this procedure is performed varying the $k$ - the number of reduced dimensions.

\begin{table*}[!htb]
    \centering
    %\begin{adjustwidth}{-0.0061in}{}
    %\begin{small}
    %\begin{tabular}{|p{0.03\textwidth}|p{0.02\textwidth}p{0.09\textwidth}|p{0.035\textwidth}p{0.035\textwidth}p{0.045\textwidth}|p{0.035\textwidth}p{0.035\textwidth}|p{0.035\textwidth}p{0.045\textwidth}|}
    \begin{tabular}{|l|ll|ccc|cc|cc|}
        \hline
        &
        \multicolumn{2}{c|}{LM} &
        \multicolumn{3}{c|}{demographics} & \multicolumn{2}{c|}{personality} & \multicolumn{2}{c|}{mental health}\\
        
        $\bm{N_{ta}}$ &
        \textbf{type} &
        \textbf{name} &
        \textbf{\begin{tabular}[c]{@{}c@{}}age\\ ($r$)\end{tabular}}  & \textbf{\begin{tabular}[c]{@{}c@{}}gen\\ ($F1$)\end{tabular}} & \textbf{\begin{tabular}[c]{@{}c@{}}gen2\\ ($F1$)\end{tabular}}& \textbf{\begin{tabular}[c]{@{}c@{}}ext\\ ($r_{dis}$)\end{tabular}}&
        \textbf{\begin{tabular}[c]{@{}c@{}}ope\\ ($r_{dis}$)\end{tabular}}&
        \textbf{\begin{tabular}[c]{@{}c@{}}bsag\\ ($r_{dis}$)\end{tabular}}& \textbf{\begin{tabular}[c]{@{}c@{}}sui\\ ($F1$)\end{tabular}}\\ 
        \hline
        \multirow{4}{*}{100}& AE & BERT & 0.533&	0.703& \textbf{0.761}&	\textbf{0.163}&	0.184&	0.424&	0.360  \\
        %\hline
         & AE & RoBERTa & \textbf{0.589}& \textbf{0.712}&	\textbf{0.761}&	0.123&	\textbf{0.203}&	\textbf{0.455}&	\textbf{0.363} \\
        %\hline
         & AR & XLNet & 0.582&	0.582&	0.744&	0.130 &	\textbf{0.203}&	0.441&	0.315 \\
        %\hline
         & AR & GPT-2 & 0.517&	0.584&	0.624&	0.082&	0.157&	0.397&	0.349 \\
        \hline
        \hline
        \multirow{4}{*}{500}& AE & BERT & 0.686&	\textbf{0.810}&	0.837&	0.278&	0.354&	0.484&	\textbf{0.466}\\
        %\hline
         & AE & RoBERTa & \textbf{0.700}&	0.802& \textbf{0.852}&	\textbf{0.283}&	\textbf{0.361}&	0.490&	0.432 \\
        %\hline
         & AR & XLNet & 0.697&	0.796&	0.821&	0.261&	0.336&	\textbf{0.508}&	0.439 \\
        %\hline
         & AR & GPT-2 & 0.646 &	0.756 &	0.762 &	0.211&	0.280 &	0.481 &	0.397 \\
        \hline

    \end{tabular}
    \caption{Comparison of most commonly used auto-encoders \textbf{(AE)} and auto-regressor \textbf{(AR)} language models after reducing the 768 dimensions to 128 using NMF and trained on 100 and 500 samples ($\bm{N_{ta}}$) for each task. ($\bm{N_{ta}}$) pertains to the number of samples used for training each task. Classification tasks (gen, gen2 and sui) were scored using macro-F1 \textbf{(F1)}; the remaining regression tasks were scored using pearson-r \textbf{(r)}/ disattenuated pearson-r ($\bm{r_{dis}}$). AE models predominantly perform the best. RoBERTa and BERT show consistent performance, with the former performing the best in most tasks. \textit{The LMs in the table were base models  (approx. 110M parameters)}.}
     \label{tab:best_lm}
    %\end{small}
    %\end{adjustwidth}
    
\end{table*}    

\subsection{Bootstrapped Sampling \& Training}
% \paragraph{Evaluation with Bootstrapped Sampling \& Training.}
We systematically evaluate the role of training sample ($N_{ta}$) versus embedding dimensions ($k$) for human-level prediction tasks. 
The approach is described in algorithm \ref{alg:method}. 
Varying $N_{ta}$, the task-specific train data (after dimension reduction) is sampled randomly (with replacement) to get ten training samples with $N_{ta}$ users each.
Small $N_{ta}$ values simulate a low-data regime and were used to understand its relationship with the least number of dimensions required to perform the best ($N_{ta}$ vs $k$). 
%  \has{move to results: $N_{ta}$ was chosen from the following values: 50, 100, 200, 500, 1000, 2000, 5000 and 10000.} 
Bootstrapped sampling was done to arrive at a conservative estimate of performance. 
Each of the bootstrapped samples was used to train either an L2 penalized (ridge) regression model or logistic regression for the regression and classification tasks respectively. The  performance on the test set using models from each bootstrapped training sample was recorded in order to derive a mean and standard error for each $N_{ta}$ and $k$ for each task.

To summarize results over the many tasks and possible $k$ and $N_{ta}$ values in a useful fashion, we propose a `\textit{first k to peak (fkp)}' metric. For each $N_{ta}$, this is the first observed $k$ value for which the mean score is within the 95\% confidence interval of the peak performance. 
This quantifies the minimum number of dimensions required for peak performance. % as a function of $N_{ta}$.

\section{Results}

\subsection{Best LM for Human-Level Tasks }
We start by comparing transformer LMs, replicating the setup of one of the state-of-the-art systems for the CLPsych-2019 task in which embeddings were reduced from BERT-base to approximately 100 dimensions using NMF~\citep{matero-etal-2019-suicide}. 
Specifically, we used 128 dimensions (to stick with powers of 2 that we use throughout this work) as we explore the other LMs over multiple tasks (we will explore other dimensions next) and otherwise use the bootstrapped evaluation described in the method.

Table \ref{tab:best_lm} shows the comparison of the four transformer LMs when varying the sample size ($N_{ta}$) between two low data regimes: 100 and 500\footnote{The performance of all transformer embeddings without any dimension reduction along with smaller sized models can be found in the appendix section \ref{appendix_LM}.}.
%The hidden state sizes of all the models was held fixed (768 per layer).
RoBERTa and BERT were the best performing models in almost all the tasks, suggesting auto-encoders based LMs are better than auto-regressive models for these human-level tasks. Further, RoBERTa performed better than BERT in the majority of cases. Since the number of model parameters are comparable, this may be attributable to RoBERTa's increased pre-training corpus, which is inclusive of more human discourse and larger vocabularies in comparison to BERT.

 \begin{table}%[!htb]
    \centering
    \begin{tabular}{|l|l|c|c|}
    %\begin{tabular}{|p{0.03\textwidth}|p{0.1\textwidth}|p{0.08\textwidth}|p{0.08\textwidth}|}
        \hline
        $N_{ta}$&
        Method&
        Age&
        Gen\\

        \hline
        \multirow{2}{*}{100} & Fine-tuned & 0.54 & 0.54 \\
        & Pre-trained & \textbf{0.56} & \textbf{0.63} \\
        \hline
        \multirow{2}{*}{500} & Fine-tuned & 0.64 & 0.60 \\
        & Pre-trained & \textbf{0.66} &  \textbf{0.74}\\
        \hline
\end{tabular}
    \caption{Comparison of task specific fine tuning of RoBERTa (top 2 layers) and pre-trained RoBERTa embeddings (second to last layer) for age and gender prediction tasks. Results are averaged across 5 trials randomly sampling users equal to $N_{ta}$ from the Facebook data and reducing messages to maximum of 20 per user.}
    \label{tab:ft}
\end{table}

\begin{table*}[!htb]
    \centering
    %\begin{adjustwidth}{-0.45in}{}
    %\begin{small}
    % \begin{tabular}{|p{0.035\textwidth}|p{0.03\textwidth}|p{0.098\textwidth}|p{0.033\textwidth}p{0.035\textwidth}p{0.045\textwidth}|p{0.035\textwidth}p{0.045\textwidth}|p{0.035\textwidth}p{0.045\textwidth}|}
    % \begin{small}
    \begin{tabular}{|l|ll|ccc|cc|cc|}
        \hline
        &
        &
        &
        \multicolumn{3}{c|}{demographics} & \multicolumn{2}{c|}{personality} &
        \multicolumn{2}{c|}{mental health} \\
        % \multicolumn{2}{|C|}{\begin{tabular}{@{}c@{}}mental\\health\end{tabular}}\\
        
        $\bm{N_{pt}}$ &
        $\bm{N_{ta}}$ &
        \textbf{Reduction} &
        \textbf{\begin{tabular}[c]{@{}c@{}}age\\ ($r$)\end{tabular}}  & \textbf{\begin{tabular}[c]{@{}c@{}}gen\\ ($F1$)\end{tabular}} & \textbf{\begin{tabular}[c]{@{}c@{}}gen2\\ ($F1$)\end{tabular}}& \textbf{\begin{tabular}[c]{@{}c@{}}ext\\ ($r_{dis}$)\end{tabular}}&
        \textbf{\begin{tabular}[c]{@{}c@{}}ope\\ ($r_{dis}$)\end{tabular}}&
        \textbf{\begin{tabular}[c]{@{}c@{}}bsag\\ ($r_{dis}$)\end{tabular}}& \textbf{\begin{tabular}[c]{@{}c@{}}sui\\ ($F1$)\end{tabular}}\\
        \hline
        \multirow{10}{*}{56k*} & \multirow{5}{*}{100}& PCA & 0.650&	\textbf{0.747}&	0.777&	\textbf{0.189}&	0.248&	\textbf{0.466}&	\textbf{0.392}  \\
        
         & & PCA-PPA & 0.517& 0.715& 0.729&	0.173&	0.176&	0.183&	0.358\\
        
         & & FA & 0.534& 0.722&	0.729&	0.171&	0.183&	0.210&	0.360 \\
        
         & & NMF & 0.589&	0.712&	0.761&	0.123&	0.203&	0.455&	0.363\\
         
         & & NLAE & \textbf{0.654}&	0.744&	\textbf{0.782}&	0.188&	\textbf{0.263}&	0.447&	0.367\\
         
        \cline{2-10}
        
        & \multirow{5}{*}{500}& PCA & \textbf{0.729}& \textbf{0.821}&	\textbf{0.856}&	\textbf{0.346}&	0.384&	\textbf{0.514}&	0.416\\
        
         & &  PCA-PPA & 0.707&	0.814&	0.849&	0.317&	0.349&	0.337&	0.415\\
        
         & & FA & 0.713& 0.819&	0.849&	0.322&	0.361&	0.400&	0.415 \\
        
         & & NMF & 0.700&	0.802&	0.852&	0.283&	0.361&	0.490&	\textbf{0.432} \\
         
         & & NLAE & 0.725&	0.820&	0.843&	0.340&	\textbf{0.394}&	0.485&	0.409\\
        
        \hline
        \hline
        
        \multirow{4}{*}{500} & \multirow{2}{*}{100}& PCA & \textbf{0.644}& \textbf{0.749}& \textbf{0.788}& \textbf{0.186}& \textbf{0.248}&	0.412& \textbf{0.392}\\
         
         & & NLAE & 0.634& 0.743&	0.744& 0.159&	0.230& \textbf{0.433}& 0.367\\
         
         \cline{2-10}
         
         & \multirow{2}{*}{500}& PCA & \textbf{0.726}& \textbf{0.819}& \textbf{0.850}& \textbf{0.344}& \textbf{0.382}&	\textbf{0.509}& \textbf{0.416}\\
         
         & & NLAE & 0.715& 0.798&	0.811& 0.312&	0.360& 0.490& 0.409\\
         
         \hline

    \end{tabular}
    % \end{small}
    %\end{adjustwidth}
    \caption{Comparison of different dimension reduction techniques of RoBERTa embeddings (penultimate layer) reduced down to 128 dimensions and $N_{ta} = $ 100 and 500. Number of user samples for pre-trianing the dimension reduction model, $\bm{N_{pt}}$ was $56k$ except for gen2, bsag (which had 9k users) and sui (which had 496 users). PCA performs the best overall and NLAE performs as good as PCA consistently. With uniform pre-training size ($N_{pt}=500$), PCA performs better than NLAE.}
    \label{tab:best_dr}
\end{table*}

\subsection{Fine-Tuning Best LM }
We next evaluate fine-tuning in these low data situations\footnote{As we are focused on readily available models, we consider substantial changes to the architecture or training as outside the scope of this systematic evaluation of existing techniques.}.
Utilizing RoBERTa, the best performing transformer from the previous experiments, we perform fine-tuning across the age and gender tasks. 
Following ~\cite{sun2019fine,mosbach2020stability}, we freeze layers 0-9 and fine-tune layers 10 and 11. Even these top 2 layers alone of RoBERTa still result in a model that is updating tens of millions of parameters while being tuned to a dataset of hundreds of users and at most 10,000 messages. 

In table \ref{tab:ft}, results for age and gender are shown for both sample sizes of 100 and 500.
For Age, the average prediction across all of a user's messages was used as the user's prediction and for gender the mode was used.   
Overall, we find that fine-tuning offers lower performance with increased overhead for both train time and modeling complexity (hyperparameter tuning, layer selection, etc).  
 
We did robustness checks for hyper-parameters to offer more confidence that this result was not simply due to the fastidious nature of fine-tuning.
The process is described in Appendix \ref{model_details}, including an extensive exploration of hyper-parameters, which never resulted in improvements over the pre-trained setup.  
We are left to conclude that fine-tuning over such small user samples, at least with current typical techniques, is not able to produce results on par with using transformers to produce pre-trained embeddings.

\subsection{Best Reduction technique for Human-Level Tasks}
We evaluated the reduction techniques in low data regime by comparing their performance on the downstream tasks across 100 and 500 training samples ($N_{ta}$).
As described in the methods, techniques including PCA, NMF and FA along with NLAE, were applied to reduce the 768 dimensional RoBERTa embeddings to 128 features.
The results in table \ref{tab:best_dr} show that PCA and NLAE perform most consistently, with PCA having the best scores in the majority tasks. NLAE's performance appears dependent on the amount of data available during the pre-training. This is evident from the results in Table \ref{tab:best_dr} where the $N_{pt}$ was set to a uniform value and tested for all the tasks with $N_{ta}$ set to 100 and 500. 
Thus, PCA appears a more reliable, showing more generalization for low samples.

\begin{figure*}[!hbt]
    \centering
    %\begin{adjustwidth}{}{}
    %\includegraphics[page=1, width=1\linewidth,trim=0cm 2cm 0cm 0cm]{assets/nvsk_noci_leg.pdf}
    \includegraphics[width=6.5in]{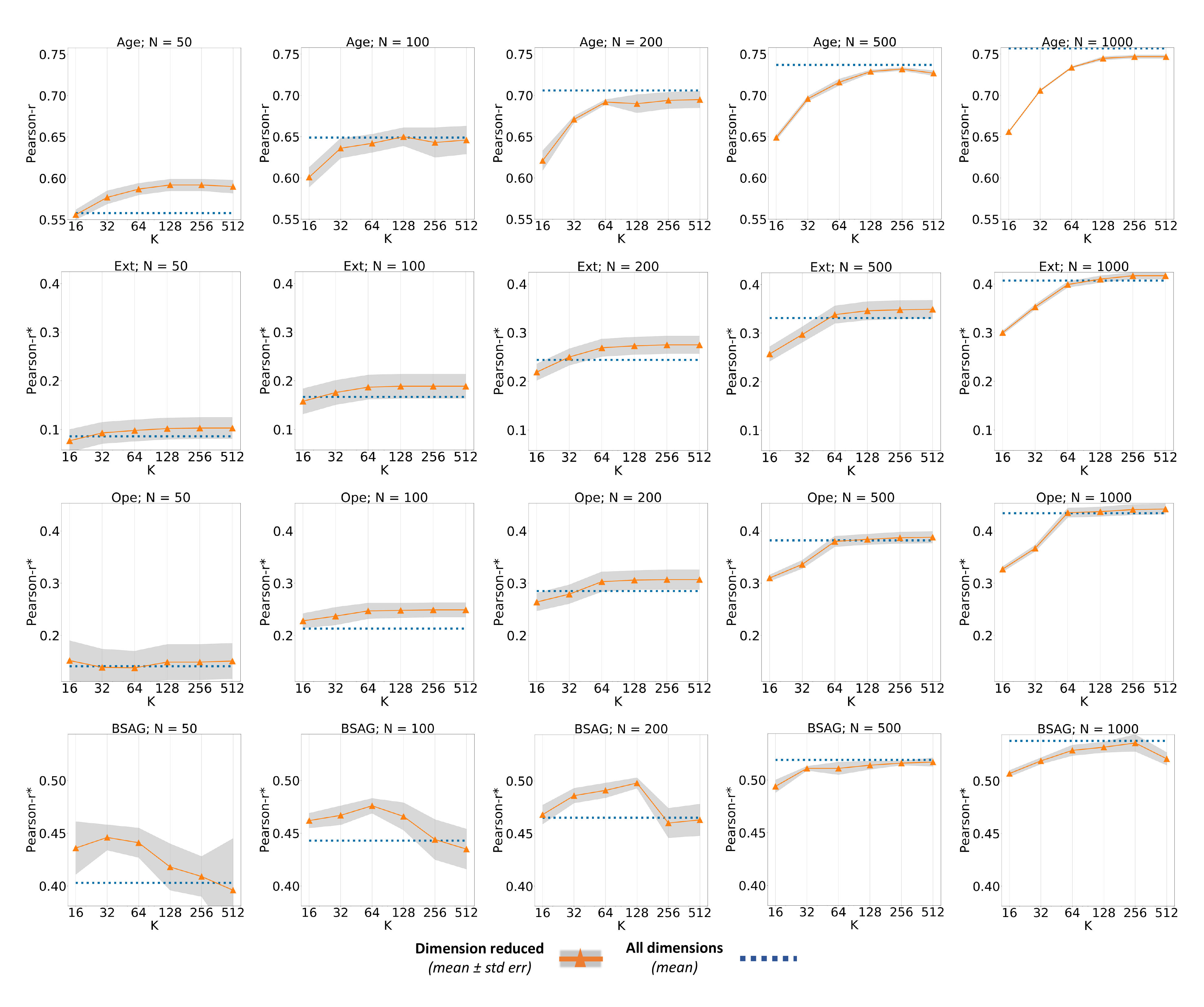}
    \caption{Comparison of performance for all regression tasks: age, ext, ope and bsag over varying $N_{ta}$ and $k$. 
    Results vary by task, but predominantly, performance at k=64 is better than the performance without any reduction. It is conclusive that the reduced features almost always performs better or as good as the original embeddings.
    %\has{add a bit more space above and below the plots}\has{legend images are still quite big relative to the plots}. 
    }
    \label{fig:KvsN_lower}
    %\end{adjustwidth}
    
\end{figure*}

\subsection{Performance by Sample Size and Dimensions}
Now that we have found (1) RoBERTa generally performed best, (2) pre-trainining worked better than fine-tuning, and (3) PCA was most consistently best for dimension reduction (often doing better than the full dimensions), we can systematically evaluate model performance as a function of training sample size ($N_{ta}$) and number of dimensions ($k$) over tasks spanning demographics, personality, and mental health. 
We exponentially increase $k$ from 16 to 512, recognizing that  variance explained decreases exponentially with dimension~\citep{mu2018all}.  
%To get a stable estimate of performance, for each sample size, we used the 10 bootstrapped training samples as previously explained. 
The performance is also compared with that of using the RoBERTa embeddings without any reduction. 

Figure \ref{fig:KvsN_lower} compares the scores at reduced dimensions for age, ext, ope and bsag. These charts depict the experiments on typical low data regime ($N_{ta} \leq 1000$). 
Lower dimensional representations performed comparable to the peak performance with just $\frac{1}{3}$ the features while covering the most number of tasks and just $\frac{1}{12}$ features for the majority of tasks. Charts exploring other ranges of $N_{ta}$ values and remaining tasks can be found in the appendix \ref{appendix_higherN_ta}.

\subsection{Least Number of Dimensions Required }

\begin{table}%[!hbt]
    \centering
    \resizebox{0.48\textwidth}{!}{
    %\begin{adjustwidth}{}{}
    \begin{tabular}{|r|c|c|c|}
        
        \hline
        $\bm{N_{ta}}$ &
        \begin{tabular}{@{}c@{}}
         \\\textbf{demographics}\\ (3 tasks)
        \end{tabular}
& 
\begin{tabular}{@{}c@{}}
         \\\textbf{personality}\\ (2 tasks)
        \end{tabular} & \begin{tabular}{@{}c@{}}
        \textbf{mental}\\ \textbf{health}\\
        (2 tasks)
        \end{tabular}\\
        
        \hline
        \textit{50}& 16& 16& 16\\
        \textit{100}& 128& 16& 22\\
        \textit{200}& 512& 32& 45\\ 
        \textit{500}& 768& 64& 64\\ 
        \textit{1000}& 768& 90& 64\\
        \hline

    \end{tabular}
    %\end{adjustwidth}
    }
    \caption{\textit{First k to peak (fkp)} for each set of tasks: the least value of k that performed statistically equivalent ($p > .05$) to the best performing setup (peak).  Integer shown is the exponential median of the set of tasks. This table summarizes comprehensive testing and \textbf{we suggest its results, \textit{fkp}, can be used as a recommendation for the number of dimensions to use given a task domain and training set size. }
    % \has{If time: can we think of some way to make this "look" more like the main result?}  
    }
    
    \label{tab:NvsK_med}
\end{table}

Lastly, we devise an experiment motivated by answering the question of how many dimensions are necessary to achieve top results, given a limited sample size. 
Specifically, we define `\textit{first k to peak}' (\textit{fkp}) as the least valued $k$ that produces an accuracy equivalent to the peak performance. 
A 95\% confidence interval was computed for the best score (peak) for each task and each $N_{ta}$ based on bootstrapped resamples, and \textit{fkp} was the least number of dimensions where this threshold was passed. 

Our goal is that such results can provide a systematic guide for making such modeling decisions in future human-level NLP tasks, where such an experiment (which relies on resampling over larger amounts of training data) is typically not feasible. 
Table \ref{tab:NvsK_med} shows the \textit{fkp} over all of the training sample sizes ($N_{ta}$). The exponential median (med) in the table is calculated as follows: $\textit{med} = 2^{\textit{Median(log(x))}}$

The \textit{fkp} results suggest that more training samples available yield ability to leverage more dimensions, but the degree to which depends on the task. In fact, utilizing all the embedding dimensions was only effective for demographic prediction tasks.
The other two tasks benefited from reduction, often with only $\frac{1}{12}$ to $\frac{1}{6}$ of the original second to last transformer layer dimensions. 

\section{Error Analysis}
Here, we seek to better understand why using pre-trained models worked better than fine-tuning, and differences between using PCA and NMF components
% , as well as differences between PCA and the full pre-trained model 
in the low sample setting ($N_{ta}=500$).

\begin{table}[!htb]
    \centering
    \begin{tabular}{|c|l|}
        \hline
         \textbf{Association}&  \textbf{LIWC variables}\\
         \hline
         Positive& \begin{tabular}{@{}l@{}}
         {Informal, Netspeak, Negemo} \\ {Swear, Anger} \end{tabular}\\
         \hline
         Negative& \begin{tabular}{@{}l@{}}
         {Affiliation, Social, We, They,}\\{Family, Function, Drives, Prep,} \\{Focuspast, Quant} \end{tabular}\\
         \hline
    \end{tabular}
    \caption{Top LIWC variables having negative and positive correlations with the difference in the absolute error of the pre-trained model and the fine-tuned model for age prediction. Benjamini-Hochberg FDR $p<.05$. This suggests that the fine-tuned models have lesser error than pre-trained model when the language is informal and consists of more affect words.}
    \label{tab:liwc_ft_full}
\end{table}

% \begin{figure}[!htb]
%     \centering
%         \includegraphics[width=0.45\linewidth]{assets/del_roba_FT_abs_neg.r_0.034-0.090.png}
%         \includegraphics[width=0.45\linewidth]{assets/del_roba_FT_abs_pos.r_0.033-0.070.png}
%     \caption{\has{just make a list of the top 5 or 10 (use multple columns to save space); NLP community generally doesn't like word clouds.} LIWC variables having negative (left) and positive correlations (right) with the difference in the absolute error of the pre-trained model and the fine-tuned model in age prediction. $p<.05$. This shows that the fine-tuned models have lesser error than pre-trained model when the language is informal and consists more affect words.}
%     \label{fig:liwc_ft_full}
% \end{figure}

\paragraph{Pre-trained vs Fine-tuned.} 
We looked at categories of language from LIWC~\cite{Tausczik2010ThePM}, correlated with the difference in the absolute error of the pre-trained and fine-tuned model in age prediction. Table \ref{tab:liwc_ft_full} suggests that pre-trained model is better at handling users with language conforming to the formal rules, and fine-tuning helps in learning better representation of the affect words and captures informal language well.
Furthermore, these LIWC variables are also known to be associated with age~\citep{schwartz2013personality}. Additional analysis comparing these two models is available in appendix \ref{pretr_v_ft_app}.

\begin{figure}
    \centering
    \includegraphics[width=0.22\textwidth]{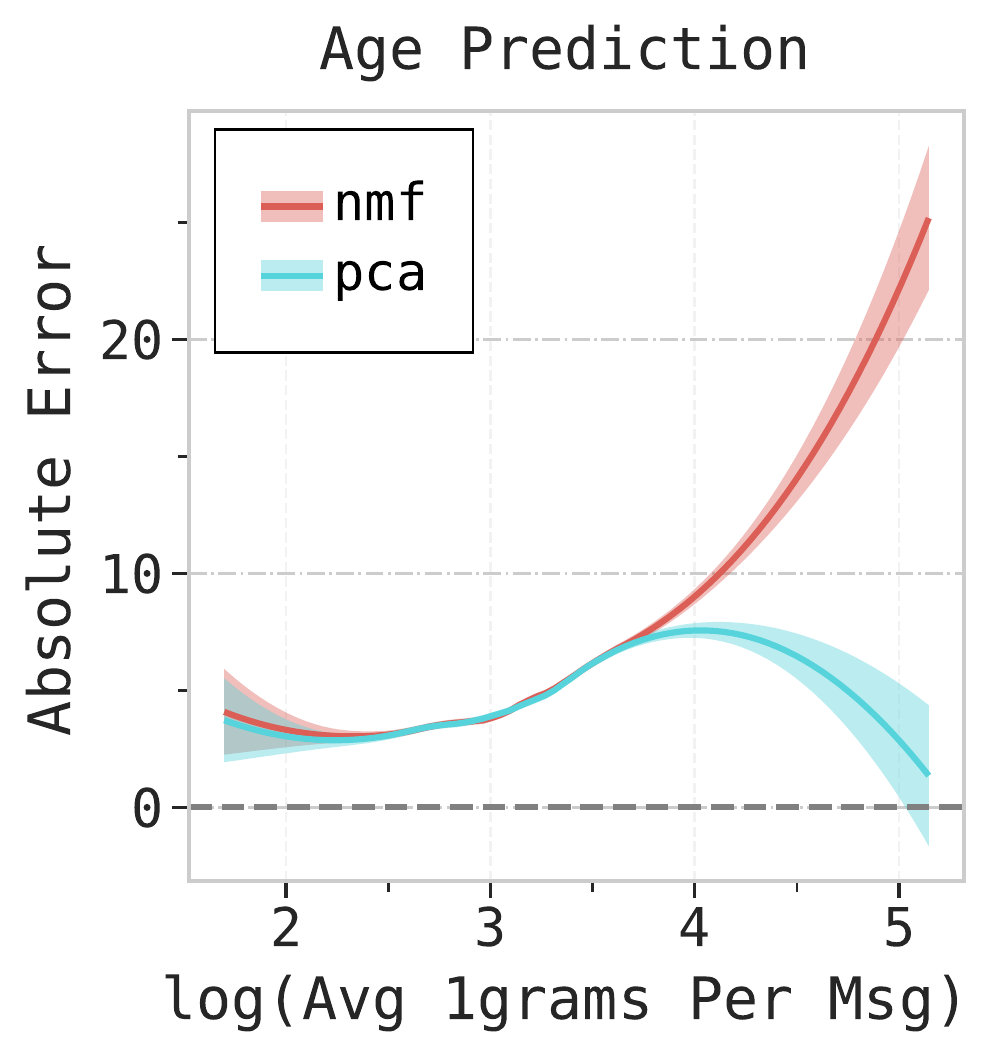}
    \includegraphics[width=0.22\textwidth]{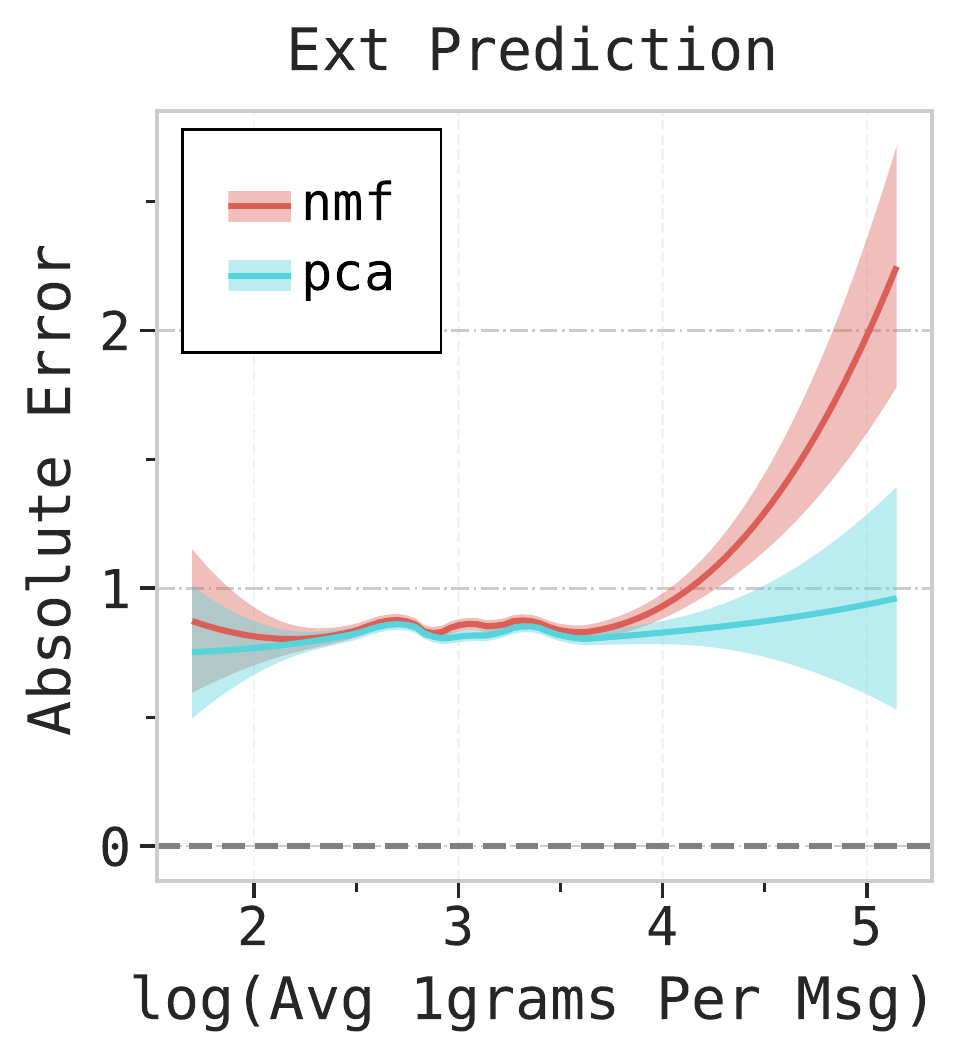}
    \caption{Comparison of the absolute error of NMF and PCA with the average number of 1 grams per message. While both the models appear to perform very similar when the texts are small or average sized, PCA is better at handling longer texts. The errors diverge when the length of the texts increases.}
    \label{fig:rpca_nmf}
\end{figure}

\paragraph{PCA vs NMF. } Figure \ref{fig:rpca_nmf} suggests that PCA is better at handling longer text sequences than NMF (> 55 one grams on avg) when trained with less data. This choice wouldn't make much difference when used for Tweet-like short texts, but the errors diverge rapidly for longer samples. We also see that PCA is better at capturing information from these texts that have higher predictive power in downstream tasks. This is discussed in appendix \ref{pca_v_nmf_app} along with other interesting findings involving the comparison of PCA and the pre-trained model in \ref{pca_v_pretr_app}.

\section{Discussion}

\paragraph{Ethical Consideration.} 
We used existing datasets that were either collected with participant consent (FB and CLPsych 2018) or public data with identifiers removed and collected in a non-intrusive manner (CLPsych 2019).
%This work has been been performed maintaining respect to the subjects' privacy in a non-intrusive way.
%To protect individual privacy, the mental health datasets were anonymized with 
All procedures were reviewed and approved by both our institutional review board as well as the IRB of the creators of the data set.  

Our work can be seen as part of the growing body of interdisciplinary research intended to understanding human attributes associated with language, aiming towards applications that can improve human life, such as producing better mental health assessments that could ultimately save lives. However, at this stage, our models are not intended to be used in practice for mental health care nor labeling of individuals publicly with mental health, personality, or demographic scores. 
Even when the point comes where such models are ready for testing in clinical settings, this should only be done with oversight from professionals in mental health care to establish the failure modes and their rates (e.g. false-positives leading to incorrect treatment or false-negatives leading to missed care; increased inaccuracies due to evolving language; disparities in failure modes by demographics).  
Malicious use possibilities for which this work is not intended include targeting advertising to individuals using language-based psychology scores, which could present harmful content to those suffering from mental health conditions.

We intend that the results of our empirical study are used to inform fellow researchers in computational linguistics and psychology on how to better utilize contextual embeddings towards the goal of improving psychological and mental health assessments. 
Mental health conditions, such as depression, are widespread and many suffering from such conditions are under-served with only 13 - 49\% receiving minimally adequate treatment ~\citep{kessler2003epidemiology, wang2005twelve}.
Marginalized populations, such as those with low income or minorities, are especially under-served~\cite{saraceno2007barriers}. Such populations are well represented in social media~\cite{pew2021} and with this technology developed largely over social media and predominantly using self-reported labels from users (i.e., rather than annotator-perceived labels that sometimes introduce bias~\cite{sap-etal-2019-risk, flekova-etal-2016-analyzing}), we do not expect that marginalized populations are more likely to hit failure modes. Still, tests for error disparities~\cite{shah-etal-2020-predictive} should be carried out in conjunction with clinical researchers before this technology is deployed. 
We believe this technology offers the potential to broaden the coverage of mental health care to such populations where resources are currently limited. 
%Mental health disorders may underlie the substantial increase in suicide mortality [94] and opioid use rates over the past decade [37].

Future assessments built on the learnings of this work, and in conjunction with clinical mental health researchers, could help the under-served by both better classifying one's condition as well as identifying an ideal treatment.  
Any applications to human subjects should consider the ethical implications, undergo human subjects review, and the predictions made by the model should not be shared with the individuals without consulting the experts.

\paragraph{Limitations. } Each dataset brings its own unique selection biases across groups of people, which is one reason we tested across many datasets covering a variety of human demographics. Most notably, the FB dataset is skewed young and is geographically focused on residents within the United States. The CLPsych 2018 dataset is a representative sample of citizens of the United Kingdom, all born on the same week, and the CLPsych-2019 dataset was further limited primarily to those posting in a suicide-related forum~\cite{zirikly-etal-2019-clpsych}. Further, tokenization techniques can also impact language model performance~\cite{bostrom-durrett-2020-byte}. 
To avoid oversimplification of complex human attributes, in line with psychological research~\cite{haslam2012categories}, all outcomes were kept in their most dimensional form -- e.g. personality scores were kept as real values rather than divided into bins and the CLPsych-2019 risk levels were kept at 4 levels to yield gradation in assessments as justified by~\citealp{zirikly-etal-2019-clpsych}.  

\section{Conclusion}
We provide the first empirical evaluation of the effectiveness of contextual embeddings as a function of dimensionality and sample size for human-level prediction tasks. Multiple human-level tasks along with many of the most popular language model techniques, were systematically evaluated in conjunction with dimension reduction techniques to derive optimal setups for low sample regimes characteristic of many human-level tasks. 

We first show the fine-tuning transformer LMs in low-data scenarios yields worse performance than pre-trained models. 
We then show that reducing dimensions of contextual embeddings can improve performance and while past work used non-negative matrix factorization~\cite{matero-etal-2019-suicide}, we note that PCA gives the most reliable improvement. 
Auto-encoder based transformer language models gave better performance, on average, than their auto-regressive contemporaries of comparable sizes. 
We find optimized versions of BERT, specifically RoBERTa, to yield the best results. 

Finally, we find that many human-level tasks can be achieved with a fraction, often $\frac{1}{6}^{th}$ or $\frac{1}{12}^{th}$, the total transformer hidden-state size without sacrificing significant accuracy.
Generally, using fewer dimensions also reduces variance in model performance, in line with traditional bias-variance trade-offs and, thus, increases the chance of generalizing to new populations. 
Further it can aid in explainability especially when considering that these dimension reduction models can be pre-trained and standardized, and thus compared across problem sets and studies.

% \section*{Acknowledgement}
% We would like to thank Youngseo Son, Mohammadzaman Zamani, and the anonymous reviewers of AAAI 2020 \& NAACL-HLT 2021 for providing helpful inputs and feedback.

\bibliography{naacl2021}
\bibliographystyle{acl_natbib}

% \clearpage

\appendix

\setcounter{table}{0}
\renewcommand{\thetable}{A\arabic{table}}
\setcounter{figure}{0}
\renewcommand{\thefigure}{A\arabic{figure}}

\section*{Appendices}
\section{Experimental Setup}
\paragraph{Implementation. }All the experiments were implemented using Python, DLATK~\citep{schwartz-etal-2017-dlatk}, HuggingFace Transformers~\cite{Wolf2019HuggingFacesTS}, and PyTorch~\citep{NEURIPS2019_9015}. The environments were instantiated with a seed value of 42, except for fine-tuning which used 1337. Code to reproduce all results is available in our github page:
\url{https://www.github.com/adithya8/ContextualEmbeddingDR/}

\paragraph{Infrastructure. } The deep learning models such as stacked-transformers and NLAE were run on single GPU with batch size given by:

\begin{equation}
    batch size = \left\lfloor{\frac{GPU\;memory - model\;size}{(floating\;precision/8) * (\delta)}}\right\rfloor
    \notag
\end{equation}
\begin{equation}
    \delta =
    \left\{
  \begin{array}{@{}l@{}}
  trainable_{params};  \hfill\text{for fine tuning}\\\\[1ex]
  (layers*hidden\_size*max\_tokens); \\
    \hfill\text{for embedding retrieval}\\[1ex]
  \end{array}
  \right.\notag
\end{equation}

where GPU memory and model sizes (space occupied by the model) are in bytes, $trainable_{params}$ corresponds to number of trainable parameters during fine tuning and $layers$ corresponds to the number of layers of embeddings required, the $hidden\_size$ is the number of dimensions in the hidden state and $max\_tokens$ is the maximum number of tokens (after tokenization) in any batch. We carried out the experiments with 1 NVIDIA Titan Xp GPU which has around 12 GB of memory.
All the other methods were implemented on CPU. 

\begin{figure}[!tbh]
    \centering
    \includegraphics[width=\linewidth,keepaspectratio]{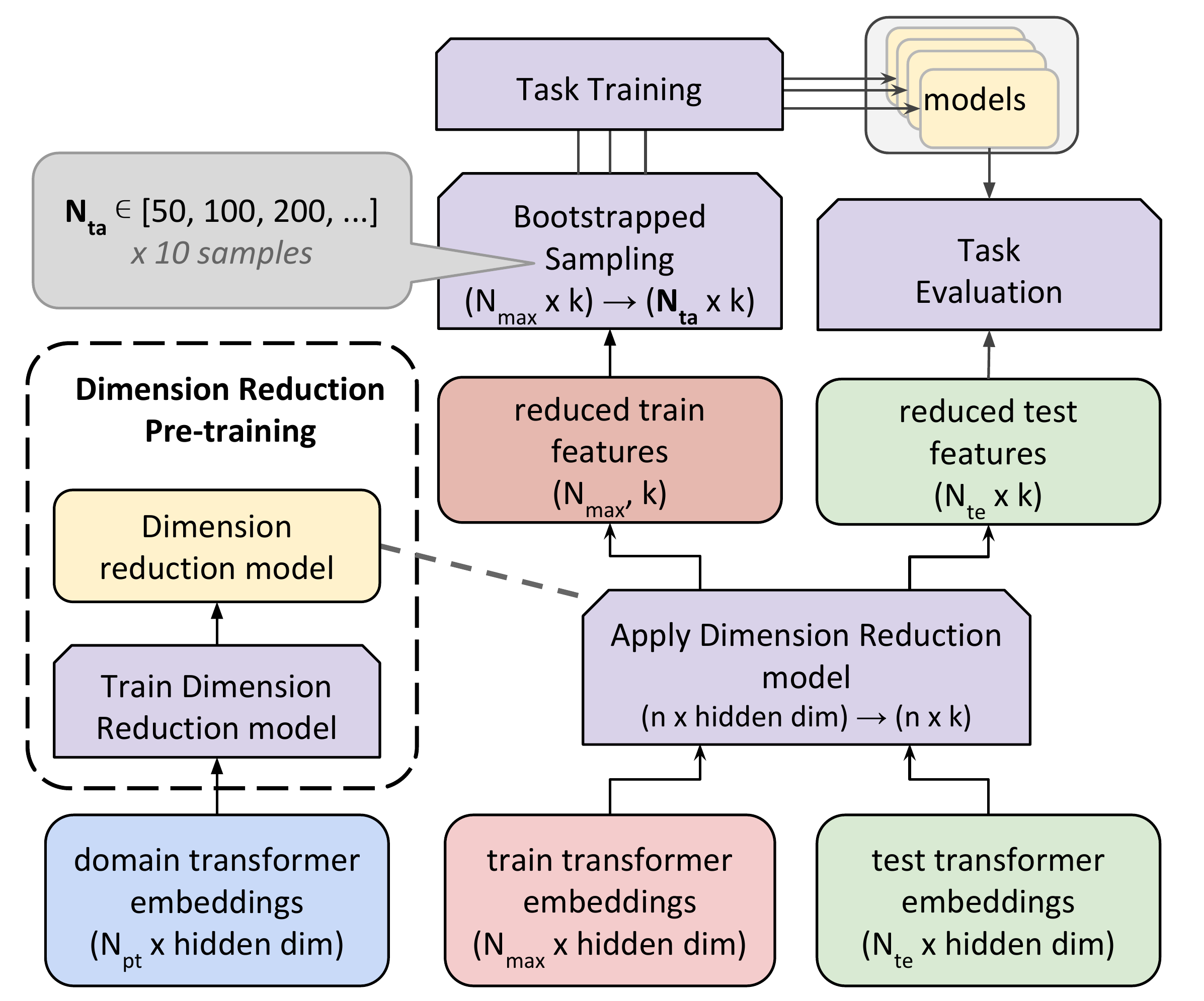}
    \caption{\textit{Depiction of Dimension Reduction method} - Transformer embeddings of domain data ($\bm{N_{pt}}$ users' embeddings\protect \footnotemark) is used to pre-train a dimension reduction model that transforms the embeddings down to $\bm{k}$ dimensions. This step is followed by applying this learned reduction model on task's train and test data embeddings. These reduced train features ($\bm{N_{max}}$ users) are then bootstrap sampled to produce 10 sets of $\bm{N_{ta}}$ users each for training task specific models. All these 10 task specific models are evaluated on the reduced test features consisting of $\bm{N_{te}}$ users during task evaluation. The mean and standard deviation of the task specific metric are collected.  
    }
    \label{fig:method}
\end{figure}

\footnotetext{Generation of user embeddings explained in detail under methods.}

\section{Model Details}
\label{model_details}
\paragraph{NLAE architecture. } The model architecture for the Non-linear auto-encoders in Table \ref{tab:best_dr} was a twin network taking inputs of 768 dimensions and reducing it to 128 dimensions through 2 layers and reconstructs the original 768 dimensional representation with 2 layers. This architecture was chosen balancing the constraints of enabling the non-linear associations while keeping total parameters low given the low sample size context. The formal definition of the model is:

\begin{gather*}
    x_{comp} = f(W_2^Tf(W_1^Tx + b_1) + b_2)\\
    x_{decomp} = \overline{W_1}^Tf(\overline{W_2}^Tx_{comp} + \overline{b_2}) + \overline{b_1}\\
    \\
    x, x_{decomp} \in R^{768}, x_{comp} \in R^{128}\\
    W_1 \in R^{768 * 448}, W_2 \in R^{448 * 128}\\ 
    b_1, \overline{b_2} \in R^{448}, b_2 \in R^{128}, \overline{b_1} \in R^{768}\\
    \overline{W_1} \in R^{448 * 768}, \overline{W_2} \in R^{128 * 448}\\
    f(a) = max(a, 0); \forall a \in R
    \label{eq:NLAE}
\end{gather*}

\paragraph{NLAE Training. } The data for domain pre-training of dimension reduction was split into 2 sets for NLAE alone: training and validation sets. 90\% of the domain data was randomly sampled for training the NLAE and the remaining 10\% of pre-training data was used to validate hyper-parameters after every epoch. This model was trained with an objective to minimise the reconstruction mean squared loss over multiple epochs. It was trained until the validation loss increased over 3 consecutive epochs. AdamW was the optimizer used with the learning rate set to 0.001. This took around 30-40 epochs depending upon the dataset.

\paragraph{Fine-tuning. }
In our fine-tuning configuration we freeze all but the top 2 layers of the best LM, to prevent over fitting and vanishing gradients at the lower layers~\cite{sun2019fine, mosbach2020stability}. 
We also apply early stopping (varied the patience between 3 and 6 depending upon the task). Other hyperparameters for this experiment include L2-regularization (in the form of weight-decay on AdamW optimizer, set to 1), dropout set to 0.3, batch size set to 10, learning rate initialized to 5e-5, and the number of epochs was set to max of 15, which was limited by early stopping between 5-10 depending on the task and early stopping patience. 

We arrived at these hyperparameter values after an extensive search. The weight decay param was searched in [100, 0.01], dropout within [0.1, 0.5], and learning rate between [5e-4, 5e-5].   

\section{Data}
Due to human subjects privacy constraints, most data are not able to be publicly distributed but they are available from the original data owners via requests for research purposes (e.g. CLPsych-2018 and CLPsych-2019 shared tasks).

\begin{table*}[htb!]
    \centering
    %\begin{adjustwidth}{-0.0061in}{}
    %\begin{small}
    \begin{tabular}{|l|ll|ccc|cc|cc|}
        \hline
        &
        \multicolumn{2}{|c|}{LM} &
        \multicolumn{3}{|c|}{demographics} & \multicolumn{2}{|c|}{personality} & \multicolumn{2}{|c|}{mental health}\\
        
        \textbf{$N_{ta}$} &
        \textbf{type} &
        \textbf{name} &
        \textbf{\begin{tabular}[c]{@{}c@{}}age\\ ($r$)\end{tabular}}  & \textbf{\begin{tabular}[c]{@{}c@{}}gen\\ ($f1$)\end{tabular}} & \textbf{\begin{tabular}[c]{@{}c@{}}gen2\\ ($f1$)\end{tabular}}& \textbf{\begin{tabular}[c]{@{}c@{}}ext\\ ($r_{dis}$)\end{tabular}}&
        \textbf{\begin{tabular}[c]{@{}c@{}}ope\\ ($r_{dis}$)\end{tabular}}&
        \textbf{\begin{tabular}[c]{@{}c@{}}bsag\\ ($r_{dis}$)\end{tabular}}& \textbf{\begin{tabular}[c]{@{}c@{}}sui\\ ($f1$)\end{tabular}}\\ 
        \hline
        \multirow{4}{*}{100}& AE & BERT & 0.615&	\textbf{0.754}&	0.758&	\textbf{0.176}&	\textbf{0.225}&	\textbf{0.457}&	\textbf{0.400}\\
        %\hline
         & AE & RoBERTa & \textbf{0.649}& 0.753&	\textbf{0.788}&	0.167&	0.213&	0.443&	0.381\\
        %\hline
         & AR & XLNet & 0.625&	0.698&	0.755&	0.144&	0.152&	\textbf{0.457}&	0.357\\
        %\hline
         & AR & GPT-2 & 0.579&	0.708&	0.681&	0.090&	0.110&	0.361&	0.335\\
        \hline
        \hline
        \multirow{4}{*}{500}& AE & BERT & 0.721&	\textbf{0.831}&	0.849&	\textbf{0.332}&	\textbf{0.395}&	0.507&	\textbf{0.489}\\
        %\hline
         & AE & RoBERTa & \textbf{0.737}&	0.830&	\textbf{0.859}&	0.331&	0.382&	\textbf{0.519}&	0.447\\
        %\hline
         & AR & XLNet & 0.715&	0.810&	0.828&	0.314&	0.364&	0.506&	0.424\\
        %\hline
         & AR & GPT-2 & 0.693&	0.794&	0.790&	0.242&	0.307&	0.508&	0.371\\
        \hline

    \end{tabular}
    %\end{small}
    %\end{adjustwidth}
    \caption{Comparison of various auto-encoders(AE) and auto-regressor(AR) language models trained on 100 and 500 samples ($N_{ta}$) for each task using all the dimensions of transformer embeddings. RoBERTa and BERT show consistent performance. }
    \label{tab:best_lm_768}
\end{table*}

\section{ Additional Results}

\subsection{ Results on higher $N_{ta}$}
\label{appendix_higherN_ta}
\begin{figure*}[!h]
    \centering
    \includegraphics[width=6in, height= 9in]{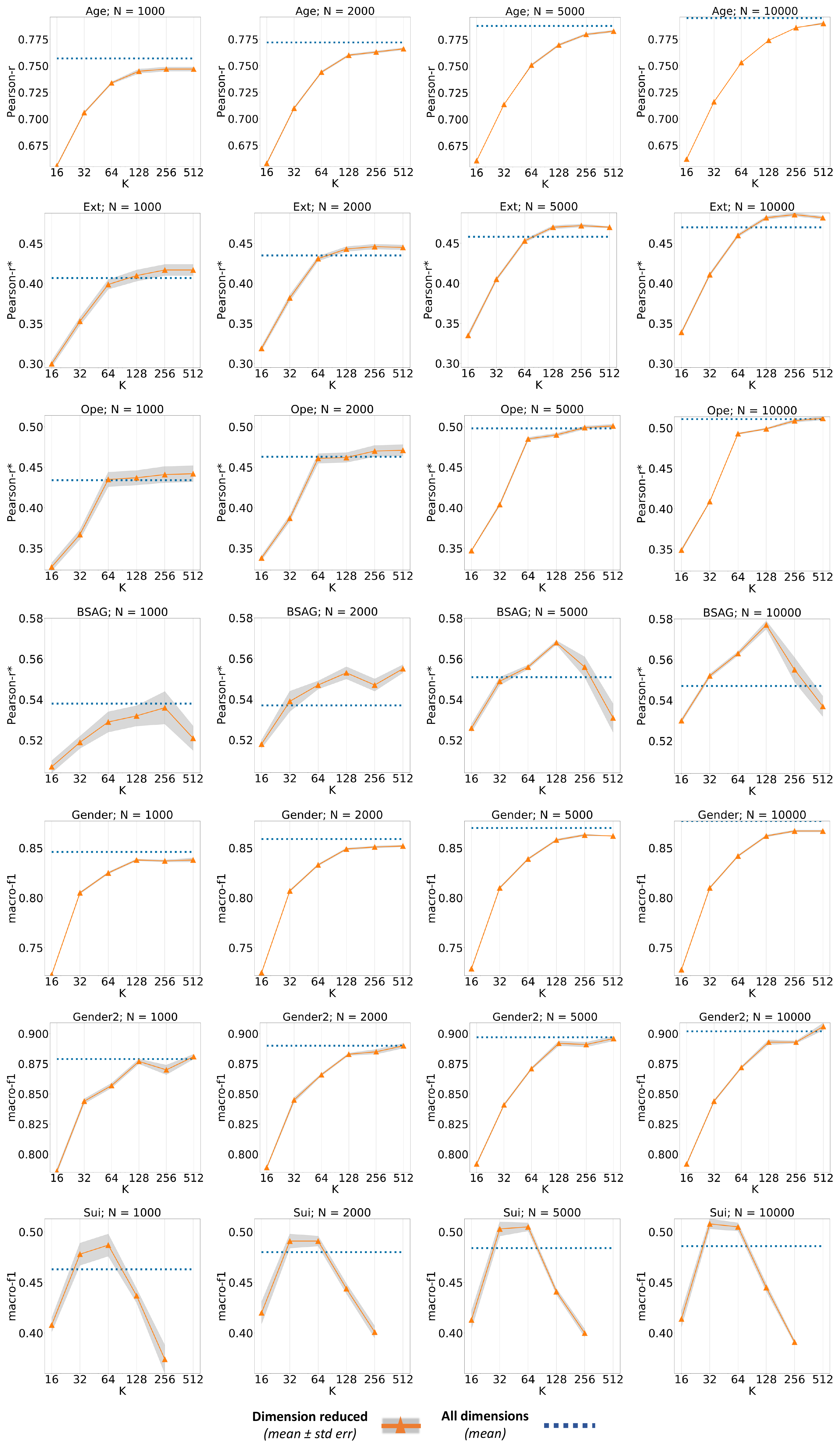}
    %\vspace*{-1cm}
    \caption{Performance recorded for reduced dimensions for all tasks at higher $N_{ta}$ values ($\geq 1000$). Reduction continues to help in performing the best in personality and mental-health tasks. The '\textit{fkp}' is observed to be shifting to a higher value, due to the rise in performance of no reduction and the reduction of standard error.}%
    \label{fig:Nvsk_higher}%
\end{figure*}

We can see that reduction still helps in majority of tasks in higher $N_{ta}$ from Figure \ref{fig:Nvsk_higher}. As expected, the performance starts to plateau at higher $N_{ta}$ values and it is visibly consistent across most tasks. With the exception of age and gender prediction using facebook data, all the other tasks benefit from reduction. 

\subsection{Results on classification tasks}
\begin{figure*}[!h]
    \centering
    \includegraphics[page=2, width=1\linewidth]{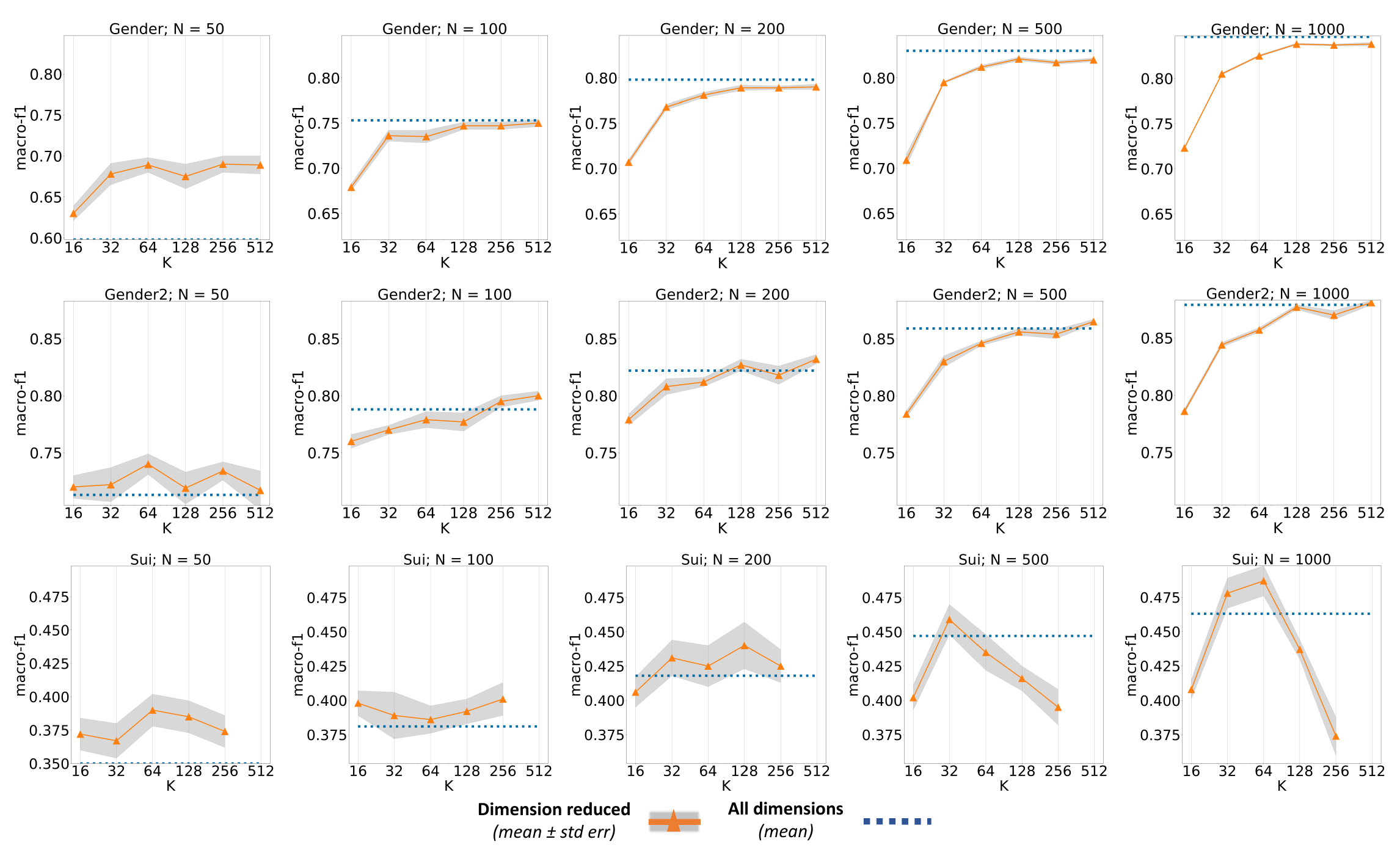}
    \caption{Comparison of performance in gen, gen2 and sui tasks for varying $N_{ta}$ between 50 and 1000.}
    \label{fig:NvsK_lower2}
\end{figure*}

Figure \ref{fig:NvsK_lower2} compares the performance of reduced dimensions at low samples size scenario ($N_{ta}\leq 1000$) in classification tasks. Except for a few $N_{ta}$ values in gender prediction using the facebook data, all the other tasks benefits from reduction in achieving the best performance.

\subsection{ LM comparison for no reduction \& Smaller models.}
\label{appendix_LM}
Table \ref{tab:best_lm_768} compares the performance of the language models without applying any dimension reduction of the embeddings and the performance of the best transformer models is also compared with smaller models (and distil version) after reducing second to last lasyer representation to 128 dimensions in table \ref{tab:best_lm_vs_small}.

\begin{table*}[!tbh]
%\begin{table*}[tbh]
    \centering
    %\begin{adjustwidth}{-0.0061in}{}
    %\begin{small}
    \begin{tabular}{|l|l|ccc|cc|cc|}
        \hline
        &
        &
        \multicolumn{3}{|c|}{demographics} & \multicolumn{2}{|c|}{personality} & \multicolumn{2}{|c|}{mental health}\\
        
        $\bm{N_{ta}}$ &
        \textbf{LM} &
        \textbf{\begin{tabular}[c]{@{}c@{}}age\\ ($r$)\end{tabular}}  & \textbf{\begin{tabular}[c]{@{}c@{}}gen\\ ($F1$)\end{tabular}} & \textbf{\begin{tabular}[c]{@{}c@{}}gen2\\ ($F1$)\end{tabular}}& \textbf{\begin{tabular}[c]{@{}c@{}}ext\\ ($r_{dis}$)\end{tabular}}&
        \textbf{\begin{tabular}[c]{@{}c@{}}ope\\ ($r_{dis}$)\end{tabular}}&
        \textbf{\begin{tabular}[c]{@{}c@{}}bsag\\ ($r_{dis}$)\end{tabular}}& \textbf{\begin{tabular}[c]{@{}c@{}}sui\\ ($F1$)\end{tabular}}\\ 
        \hline
        \multirow{4}{*}{100}&  BERT & 0.533&	0.703& \textbf{0.761}&	\textbf{0.163}&	0.184&	0.424&	0.360  \\
        %\hline
         &  RoBERTa & \textbf{0.589}& \textbf{0.712}&	\textbf{0.761}&	0.123&	0.203&	\textbf{0.455}&	\textbf{0.363} \\
        %\hline
         & DistilRoBERTa & 0.568&	0.640&	0.731&	0.130&	\textbf{0.207}&	0.446&	0.355 \\
         & ALBERT & 0.525&	0.689&	0.710&	0.111&	0.218&	0.413&	 0.355\\
        \hline
        \hline
        \multirow{4}{*}{500}& BERT & 0.686&	\textbf{0.810}&	0.837&	0.278&	0.354&	0.484&	\textbf{0.466}\\
        %\hline
         & RoBERTa & \textbf{0.700}&	0.802& \textbf{0.852}&	\textbf{0.283}&	\textbf{0.361}&	0.490&	0.432 \\
        %\hline
         & DistilRoBERTa & 0.687&	0.796&	0.826&	0.246&	0.346&	0.503&	 0.410\\
         & ALBERT & 0.668&	0.792&	0.799&	0.237&	0.337&	0.453&	 0.385\\
        \hline

    \end{tabular}
    \caption{Comparison of the best performing auto-encoder models with a smaller LMs (like ALBERT~\citep{lan2019albert} and DistilRoBERTa~\cite{sanh2019distilbert} after reduction to 128 dimensions. These results suggest that the reduction of the larger counterparts produce better results than reducing these smaller LMs' representations.}
     \label{tab:best_lm_vs_small}
    %\end{small}
    %\end{adjustwidth}
    
%\end{table*}    
\end{table*}

\subsection{Least dimensions required: Higher $N_{ta}$}

The '\textit{fkp}' plateaus as the the number of training samples grow as seen in table \ref{tab:NvsK_med_highN}. 

\begin{table}[b]
    \centering
    %\begin{adjustwidth}{}{}
    \begin{tabular}{|r|c|c|c|}
        
        \hline
        $\bm{N_{ta}}$ &
        \textbf{demographics} & \textbf{personality} & \begin{tabular}{@{}c@{}}
        \textbf{mental}\\ \textbf{health}
        \end{tabular}\\
        
        \hline
        \textit{2000}& 768& 90& 64\\
        \textit{5000}& 768& 181& 64\\
        \textit{10000}& 768& 181& 64\\
        \hline

    \end{tabular}
    %\end{adjustwidth}
    \caption{\textit{First k to peak} for each set of tasks: the least value of k that performed statistically equivalent ($p > .05$) to the best performing setup (peak). Integer shown is the exponential median of the set of tasks.}
    \label{tab:NvsK_med_highN}
\end{table}

\begin{figure}
    \centering
    \includegraphics[height=1.7in]{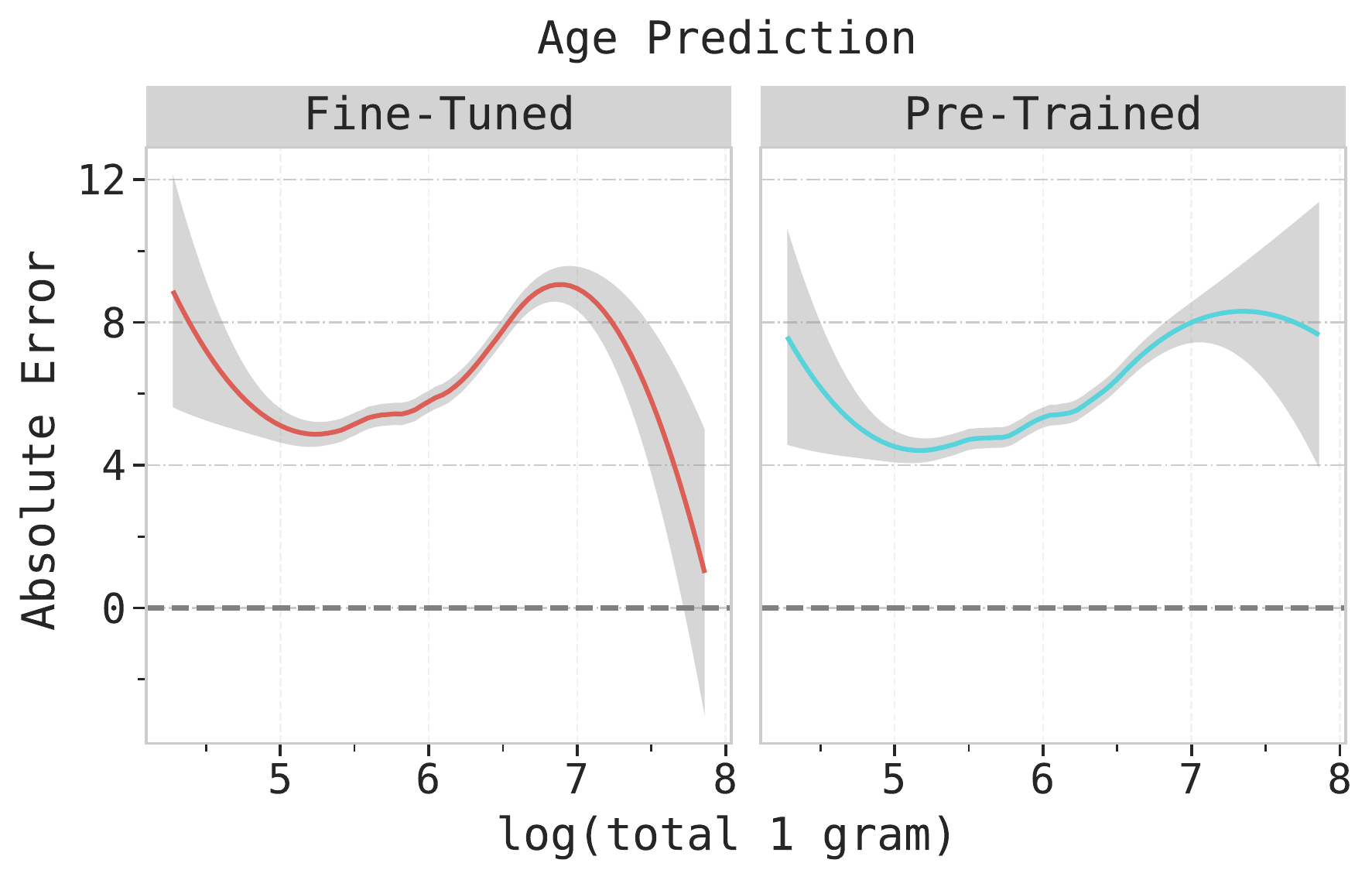}
    \caption{The absolute error in age prediction for the fine-tuned model is higher than pre-trained models for users with short messages. Fine-tuned models have smaller errors for users with longer messages.}
    \label{fig:ft_vs_pretr}
\end{figure}

\section{Additional Analysis}

\subsection{Pre-trained vs Fine-Tuned models} 
\label{pretr_v_ft_app}
We also find that fine-tuned model doesn't perform better than the pre-trained model for users with typical message lengths, but is better at handling longer sequences upon training it on the tasks' data. This is evident from the graphs in figure \ref{fig:ft_vs_pretr}.

\subsection{PCA vs NMF. } 
\label{pca_v_nmf_app}
\begin{figure}
    \centering
    \includegraphics[width=0.18\textwidth]{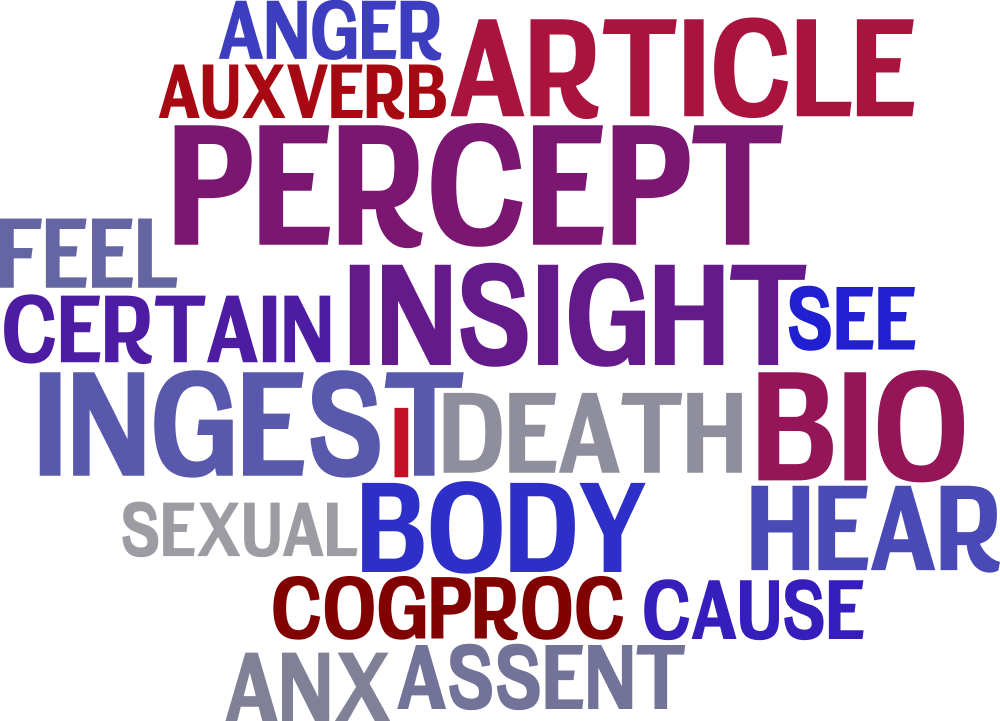}
    \includegraphics[width=0.26\textwidth]{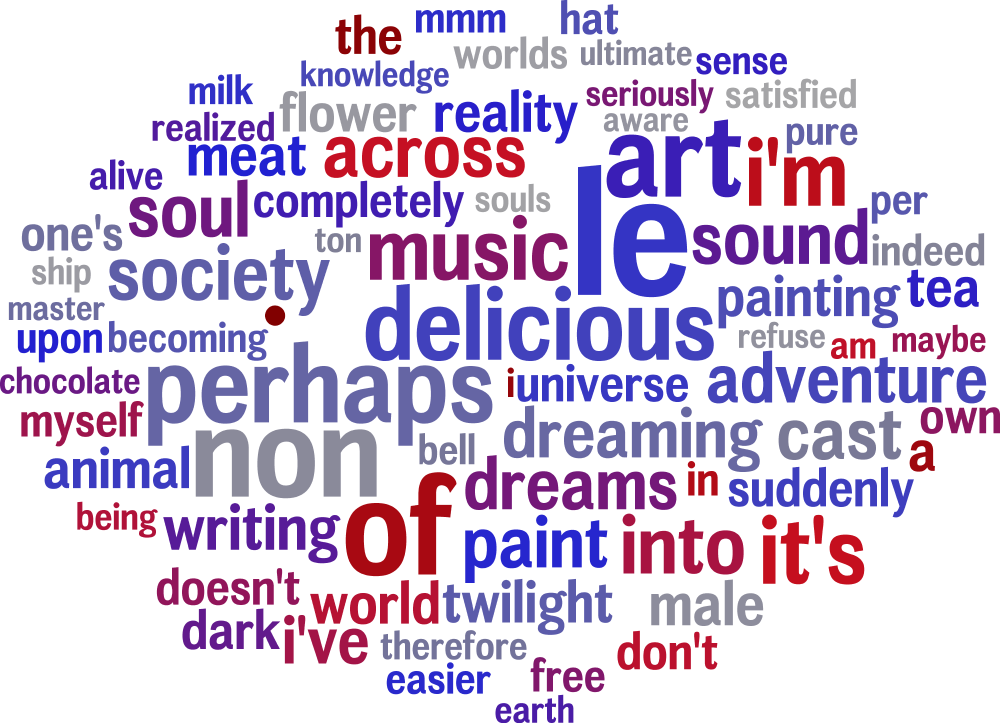}
    \caption{The word cloud of the LIWC variables (left) and the 1 grams (right) having negative correlation with the difference in the absolute error of PCA and NMF in Openness prediction. Benjamini-Hochberg FDR. $p<.05$. We can see that LIWC variables and 1 grams more correlative of a person exhibiting more openness are better captured by the PCA model than the NMF.}
    \label{fig:pca_v_nmf_app}
\end{figure}

From figure \ref{fig:pca_v_nmf_app}, we can see that LIWC variables like ARTICLE, INSIGHT, PERCEPT (perceptual process), COGPROC (cognitive process) negatively correlates to the difference in absolute error of PCA and NMF. These variables also happen to have higher correlation with the openness scores~\citep{schwartz2013personality}. We also see that characteristics typical of an open person like interest in arts, music, and writing~\citep{kern2014online} appear in the word clouds. 

\begin{figure}
    \centering
    \includegraphics[width=.22\textwidth]{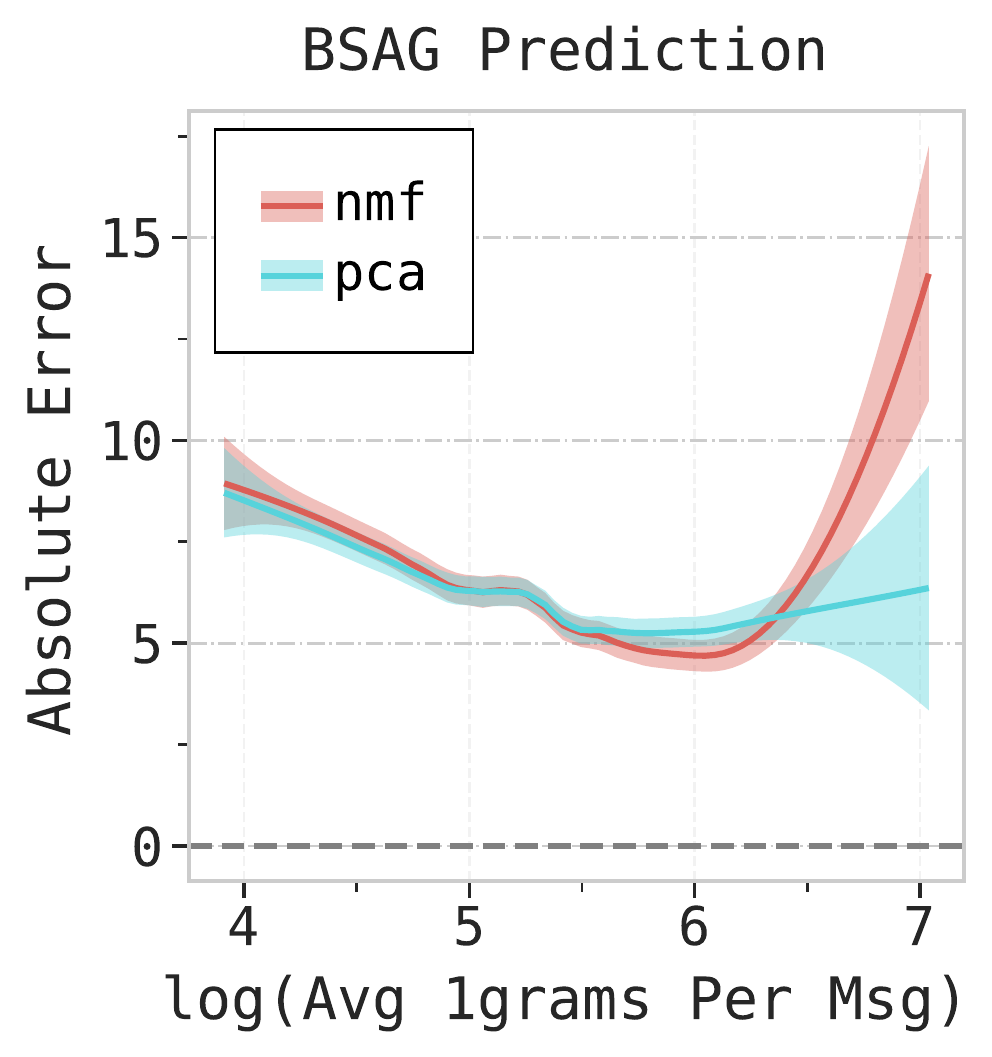}
    \includegraphics[width=.22\textwidth]{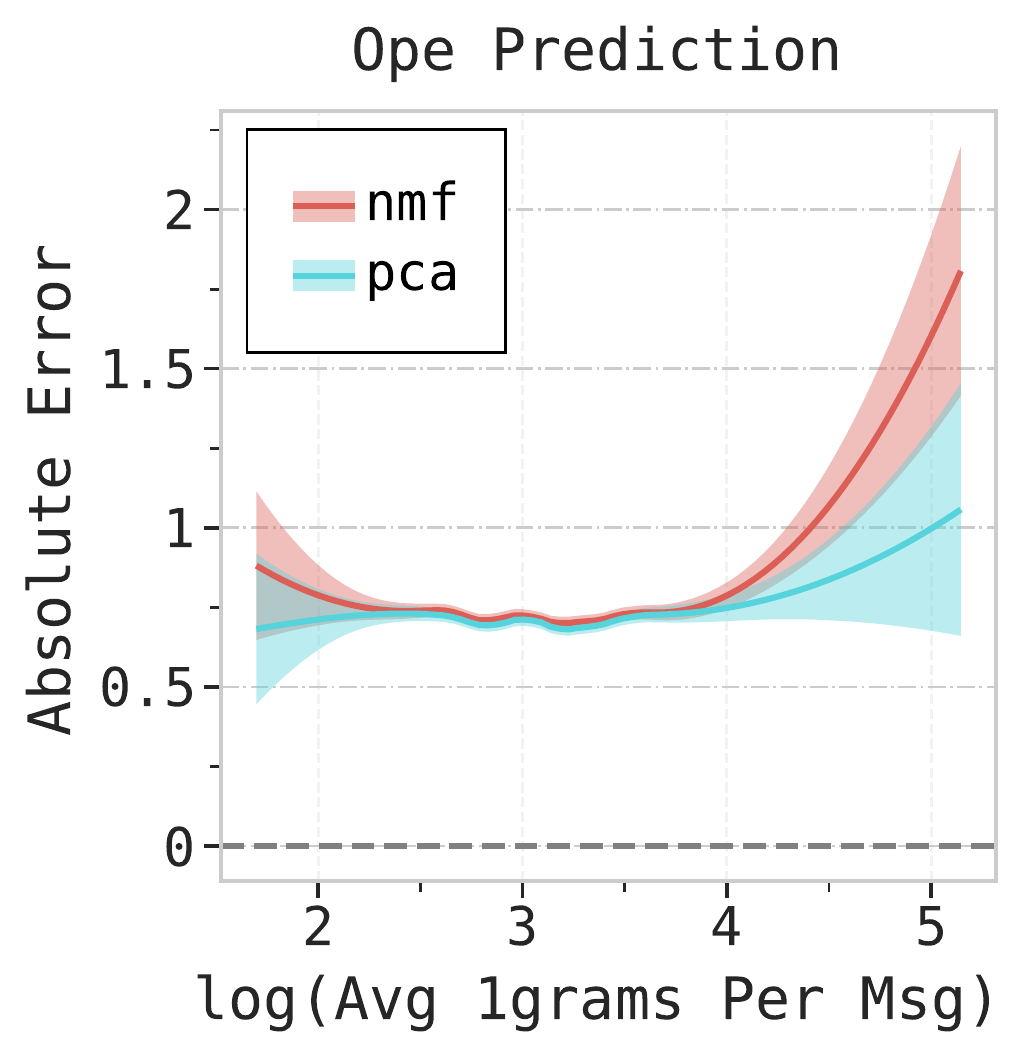}
    \caption{Comparison of the absolute error of NMF and PCA with the average number of 1 grams per message. We see that the absolute error of NMF models starts diverging at longer text sequences for the bsag and the ope tasks as well.}
    \label{fig:pca_v_nmf_avg1grm_app}
\end{figure}

The divergence of the absolute errors in NMF and PCA is seen in bsag and ope tasks as well. From graphs in figure \ref{fig:pca_v_nmf_avg1grm_app} we can see that the sequence length at which we see this behavior is close to the previously observed value in age and ext tasks.

\begin{figure}
    \centering
    \includegraphics[width=0.22\textwidth]{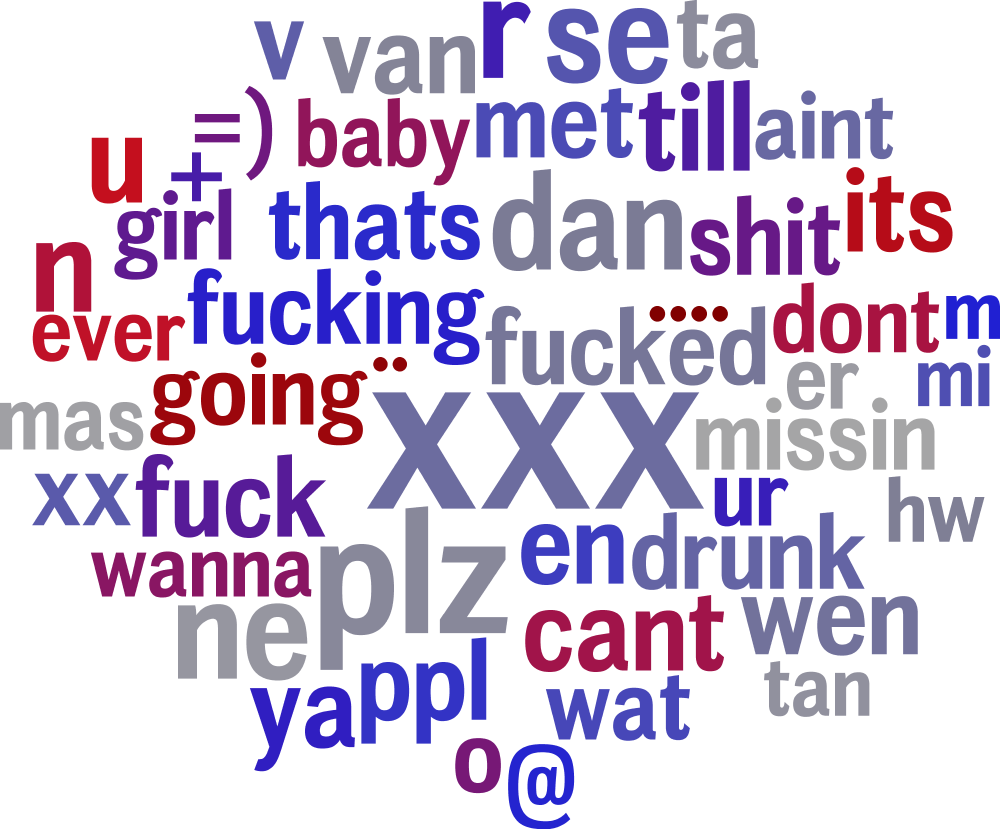}
    \includegraphics[width=0.22\textwidth]{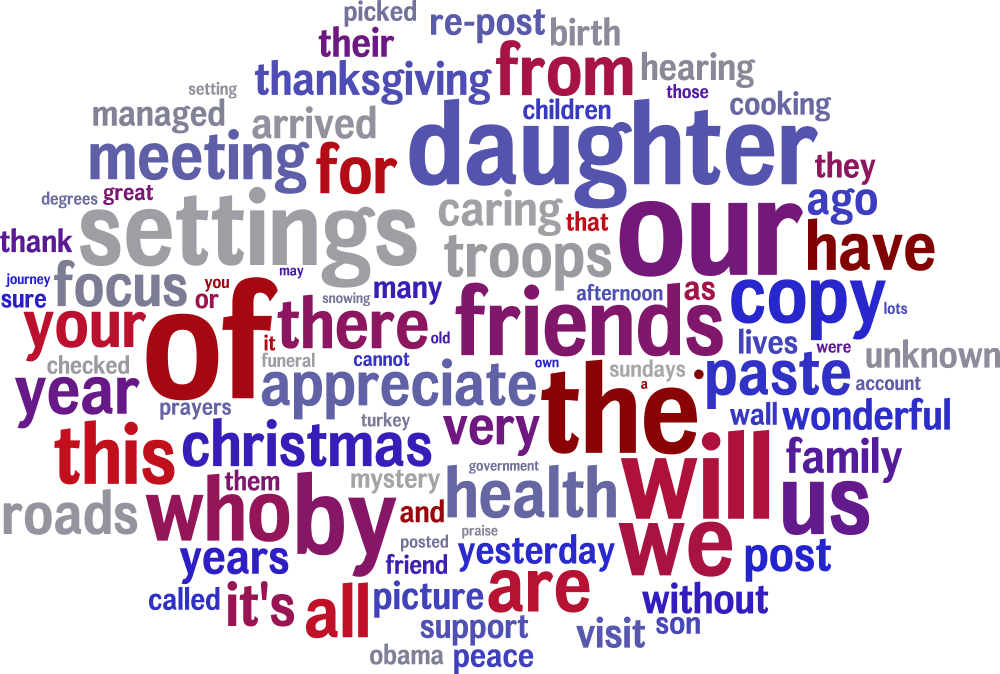}
    \caption{
    % \has{consider making list of top 20 terms instead of cloud; might be ok -- only consider if time.}.
    Terms having negative (left) and positive (right) correlations with the difference in the absolute error of the PCA and pre-trained model in age prediction. Benjamini-Hochberg FDR. $p<.05$. The error in the PCA model is lesser than pre-trained models when messages contain more slang, affect words and social media abbreviations.}
    \label{fig:pca_vs_pretr}
\end{figure}

\subsection{PCA vs Pre-trained.} 
\label{pca_v_pretr_app}
PCA models overall perform better than pre-trained model in low sample regime and from figure \ref{fig:pca_vs_pretr}, we can see that PCA captures slang, affect and standard social media abbreviations better than the pre-trained models. The task specific linear layer is better able to capture social media language with fewer dimensions (reduced by PCA) than from the original 768 features produced by the pre-trained models.

\end{document}